\documentclass[journal,compsoc]{IEEEtran}

\usepackage[dvipsnames,table,xcdraw]{xcolor}
\usepackage{booktabs}
\usepackage{graphicx}
\usepackage{amsmath}
\usepackage{amssymb}
\usepackage{multirow}
\usepackage{bm}
\usepackage[ruled]{algorithm}
\usepackage{algpseudocode}
\usepackage{overpic}
\usepackage{adjustbox}
\usepackage[caption=false]{subfig}
\usepackage[breaklinks=true,colorlinks,citecolor=black,linkcolor=black,urlcolor=black]{hyperref}

\usepackage{silence}
\usepackage{pifont}
\newcommand{\figref}[1]{Fig.~\ref{#1}}
\newcommand{\tabref}[1]{Tab.~\ref{#1}}

\newcommand{\secref}[1]{Sec.~\ref{#1}}
\newcommand{\myPara}[1]{\vspace{2pt}\noindent\textbf{#1}}

\newcommand{\sArt}{state-of-the-art}

\graphicspath{{./figs/}}

\usepackage{makecell}
\newcommand{\highlight}[1]{\textbf{\textcolor{black}{#1}}}

\def\eg{\emph{e.g.,~}}

\renewcommand{\eqref}[1]{Eq.~(\ref{#1})}

\newcommand{\namesrformer}{SRFormer}
\newcommand{\nameofmethod}{SRFormerV2}
\newcommand{\tablestyle}[2]{\small\setlength{\tabcolsep}{#1}\renewcommand{\arraystretch}{#2}\centering}

\begin{document}

\title{\nameofmethod{}: Taking a Closer Look at Permuted Self-Attention for Image Super-Resolution}

\author{Yupeng Zhou,
        Zhen Li,
        Chun-Le Guo,
        Li Liu,~\IEEEmembership{Senior Member,~IEEE},
        Ming-Ming Cheng,~\IEEEmembership{Senior Member,~IEEE},
        Qibin Hou,~\IEEEmembership{Member,~IEEE}
\IEEEcompsocitemizethanks{
\IEEEcompsocthanksitem Y. Zhou, Z. Li, C.L. Guo, M.M. Cheng, and Q. Hou are with VCIP, CS, Nankai University, Tianjin, China. 
\IEEEcompsocthanksitem L. Liu is with NUDT, Changsha, China.
\IEEEcompsocthanksitem Q. Hou is the corresponding author.
\IEEEcompsocthanksitem The previous version of this paper has been accepted by ICCV 2023~\cite{zhou2023srformer}.
}

\thanks{Manuscript received March 1, 2022; revised August 26, 2022.}}
%
%

\markboth{Journal of \LaTeX\ Class Files,~Vol.~14, No.~8, August~2015}%
{Shell \MakeLowercase{\textit{et al.}}: Bare Demo of IEEEtran.cls for Computer Society Journals}

\IEEEtitleabstractindextext{%
\begin{abstract}
Previous works have shown that increasing the window size for Transformer-based image super-resolution models (e.g., SwinIR) can significantly improve the model performance. 
Still, the computation overhead is also considerable when the window size gradually increases.
In this paper, we present \namesrformer, a simple but novel method that can enjoy the benefit of large window self-attention but introduces even less computational burden.
The core of our \namesrformer{}  is the permuted self-attention (PSA), which strikes an appropriate balance between the channel and spatial information for self-attention.
Without any bells and whistles, we show that our \namesrformer{} achieves
a 33.86dB PSNR score on the Urban100 dataset, which is 0.46dB higher than that of
SwinIR but uses fewer parameters and computations.
In addition, we also attempt to scale up the model by further enlarging the window size and channel numbers to explore the potential of Transformer-based models.
Experiments show that our scaled model, named \nameofmethod, can further improve the results and achieves \sArt.
We hope our simple and effective approach could be useful for future research in super-resolution model design. The homepage is \url{ https://z-yupeng.github.io/SRFormer/}.
\end{abstract}

}

\maketitle

\IEEEdisplaynontitleabstractindextext

\IEEEpeerreviewmaketitle

\IEEEraisesectionheading{\section{Introduction}\label{sec:introduction}}

\IEEEPARstart{S}{ingle} image super-resolution (SR) aims to restore a high-quality image from its degraded low-resolution version.
Exploring efficient and effective super-resolution algorithms has been a hot research topic in computer vision, which has a variety of applications~\cite{jo2018deep,wang2019edvr,anwar2020deep}. 
Since the pioneer works~\cite{dong2014learning,kim2016accurate,zhang2018learning,ledig2017photo,shi2016real,lim2017enhanced}, CNN-based methods have been mainstream for image super-resolution for a long time.
These methods mostly take advantage of residual learning~\cite{tai2017image,ledig2017photo,lim2017enhanced,kim2016accurate,zhang2021plug,li2018multi}, dense connections~\cite{wang2018esrgan,zhang2018residual,tong2017image}, or channel attention~\cite{zhang2018image,yang2021image} to construct network architectures, making great contributions to the development of super-resolution models.

Despite the success made by CNN-based models in super-resolution, recent works~\cite{chen2021pre,liang2021swinir,zhang2022swinfir,zhang2022efficient}
have shown that Transformer-based models perform better.
They observe that self-attention, which can build pairwise relationships, is a more efficient way to produce high-quality super-resolution images than convolutions. 
One typical work among them should be SwinIR~\cite{liang2021swinir}, which introduces Swin Transformer~\cite{liu2021swin} to image super-resolution, greatly improving the CNN-based models on various benchmarks.
Later, a variety of works, such as SwinFIR~\cite{zhang2022swinfir}, ELAN~\cite{zhang2022efficient}, and HAT~\cite{chen2022activating}, further develop SwinIR and use Transformers to design different network architectures for SR.

\begin{figure}
    \centering
    \scriptsize
    \includegraphics[width=\linewidth]{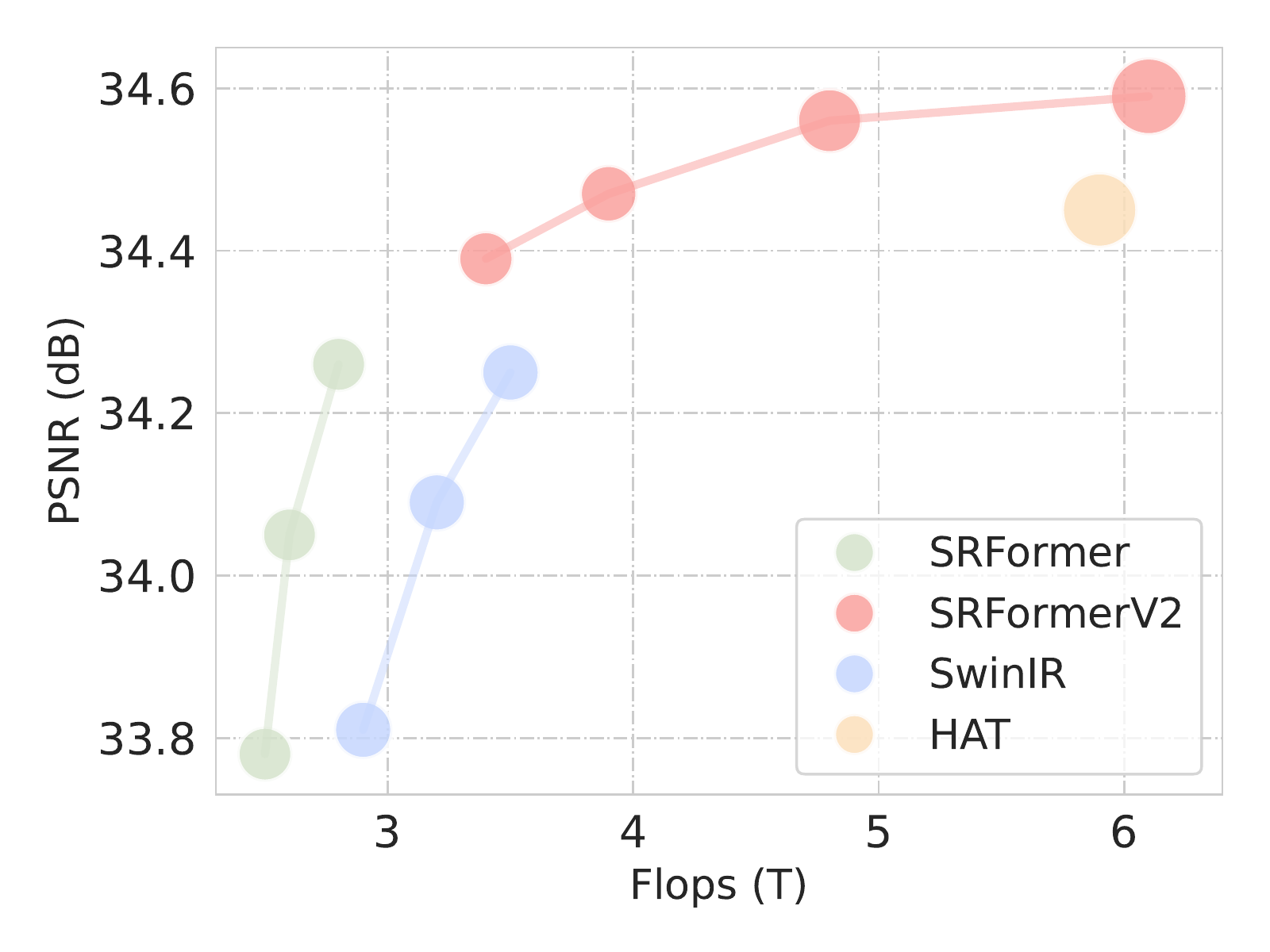}
    \put(-209, 53){$\mathrm{WS}:12$}
    \put(-199, 125){$\mathrm{WS}:24$}
    \put(-207, 95){$\mathrm{WS}:16$}
     \put(-47, 132){$\mathrm{WS}:16$}
     \put(-150, 133){$\mathrm{WS}:36$}
     \put(-125, 147){$\mathrm{WS}:36$}
     \put(-90, 153){$\mathrm{WS}:36$}
     \put(-40, 157){$\mathrm{WS}:36$}
    \put(-172, 42){$\mathrm{WS}: 8$}
    \put(-157, 85){$\mathrm{WS}:12$}
    \put(-142, 114){$\mathrm{WS}: 16$}
    \caption{Performance comparison among SwinIR, HAT, our \namesrformer{} and \nameofmethod. 
      The radius of the circles represents the parameters of the models.  ``WS'' stands for the attention window size, e.g. ``WS: 24 '' stands for  $24\times24$ attention windows.
      Our \namesrformer{} enjoys large window sizes with even fewer computations but higher PSNR scores.
    }\label{fig:teaser}
\end{figure}
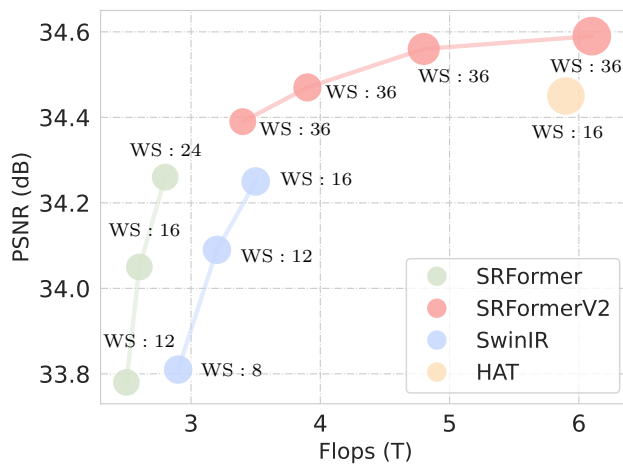

The methods above reveal that properly enlarging the windows for the shifted window self-attention in SwinIR can result in clear performance gain (see \figref{fig:teaser}).
However, the computational burden is also an important issue as the window size becomes larger. 
In addition, Transformer-based methods utilize self-attention and require networks of larger channel numbers than previous CNN-based methods~\cite{zhang2018image,zhang2018residual,kim2016accurate}.
To explore efficient and effective super-resolution algorithms, a straightforward question should be: How would the performance go if we reduce the channel number and increase the window size?

Motivated by the above question, in our conference version of this work \cite{zhou2023srformer}, we present permuted self-attention (PSA), an efficient way to build pairwise relationships within large windows (\eg $24\times24$).
The intention is to enable more pixels to participate in the attention map computation while introducing no extra computational burden.
To this end, we propose to shrink the channel dimensions of the key and value matrices and adopt a permutation operation to convey part of the spatial information into the channel dimension.
In this way, despite the channel reduction, there is no loss of spatial information, and each attention head is also allowed to keep a proper number of channels to produce expressive attention maps~\cite{touvron2021going}.
In addition, we also improve the original feed-forward network (FFN) by
adding a depth-wise convolution between the two linear layers, which we found
helps in high-frequency component recovery.

Given the proposed PSA, we construct a novel network for SR in the conference version, termed \namesrformer.
%
%
Benefiting from the proposed PSA, our \namesrformer{} receives good performance on five widely-used datasets.
Notably, for $\times2$ SR, our \namesrformer{} trained on only the DIV2K dataset~\cite{lim2017enhanced} achieves a 33.86 PSNR score on the challenging Urban100 dataset~\cite{huang2015single}.
This result is much higher than the popular SwinIR (33.40) and ELAN (33.44).
A similar phenomenon can be observed when evaluating the $\times3$ and $\times4$ SR tasks.
In addition, we perform experiments using a light version of our \namesrformer{}.
Our method performs better on all benchmarks than previous lightweight SR models.

Despite the good performance of our \namesrformer{}, the scaling ability of \namesrformer{} has not been explored well.
In this paper, we intend to further investigate the scaling ability of \namesrformer{} and conduct a series of scaling operations. 
To our knowledge, due to PSA's efficient computation of window attention, we are the first to implement self-attention computation within $40\times40$ windows.
Experimental results demonstrate that conducting attention calculations within larger windows can further improve performance, showing our method's excellent scaling capabilities.
By conducting a series of scaling operations, our new model, named \nameofmethod, is able to surpass the \sArt~HAT method~\cite{chen2022activating} with less computational cost and parameters.
To sum up, the contributions of this paper can be summarized as follows:
\begin{itemize}
    \item We propose a novel permuted self-attention mechanism for image super-resolution, which can enjoy large-window self-attention by transferring spatial information into the channel dimension. By leveraging it, we are able to implement the self-attention mechanism within $40\times40$ windows at an acceptable time complexity for SR.
    \item We build a new transformer-based super-resolution network, dubbed \namesrformer{}, based on the proposed PSA and an improved FFN from the frequency perspective (ConvFFN). 
    \item Based on \namesrformer{}, we further explore the scaling capability of the model at the macro structural level and consider the use of both large and small window PSA. The outcome of the above investigations is the new \nameofmethod{}, which is able to surpass previous \sArt~models using fewer parameters and computational cost.
\end{itemize}

\begin{figure*}
    \centering
    \small
    \setlength{\abovecaptionskip}{5pt}
    \includegraphics[width=\linewidth]{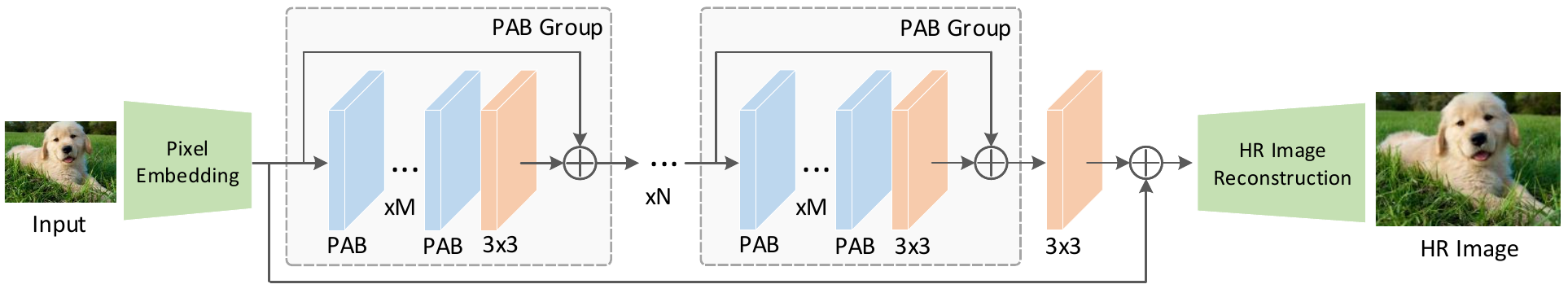}
    \caption{Overall architecture of \namesrformer{}. The pixel embedding module is a $3\times3$ convolution to map the input image to feature space. The HR image reconstruction module contains a $3\times3$ convolution and a pixel shuffle operation to reconstruct the high-resolution image. The middle feature encoding part has $N$ PAB groups, followed by a $3\times3$ convolution.}
    \label{fig:arch}
   \vspace{-4mm}
\end{figure*}
\section{Related Work}

In this section, we briefly review the literature on image super-resolution.
We first introduce the CNN-based methods and then describe the recently popular Transformer-based models.

\subsection{CNN-Based Image Super-Resolution}

Since SRCNN\cite{dong2014learning} first introduced CNN into image super-resolution (SR), many CNN-based  SR models have emerged.
DRCN~\cite{kim2016deeply} and DRRN~\cite{tai2017image} introduce recursive convolutional networks to increase the depth of the network without increasing the parameters.
Some early CNN-based methods~\cite{tai2017memnet,dong2014learning,kim2016deeply,tai2017image} attempt to interpolate the low-resolution (LR) as input, which results in a computationally expensive feature extraction.
To accelerate the SR inference process, FSRCNN~\cite{dong2016accelerating} extracts features at the LR scale and conducts an upsampling operation at the end of the network.
This pipeline with  pixel shuffle upsampling~\cite{shi2016real} has been widely used in later works~\cite{zhang2022efficient,zhang2018image,liang2021swinir}.
LapSRN~\cite{lai2017deep} and DBPN~\cite{haris2018deep} perform upsampling during extracting features to learn the correlation between LR and HR.
%
%
There are also some works~\cite{ledig2017photo,wang2018esrgan,zhang2019ranksrgan,wang2021real} that use GAN~\cite{goodfellow2014generative} to generate realistic textures in reconstruction.
MemNet~\cite{tai2017memnet}, RDN~\cite{zhang2018residual}, and HAN~\cite{niu2020single}  efficiently aggregate the intermediate features to enhance the quality of the reconstructed images.
Non-local attention~\cite{wang2018non} has also been extensively explored in SR to better model the long-range dependencies.
Methods of this type include CS-NL~\cite{mei2020image}, NLSA~\cite{mei2021image}, SAN~\cite{dai2019second}, IGNN~\cite{zhou2020cross}, etc.

\subsection{Vision Transformers}

Transformers have recently shown great potential in a variety of vision tasks,
including image classification~\cite{dosovitskiy2020image,touvron2020training,Yuan_2021_ICCV,wang2021pyramid,yuan2022volo}, object detection~\cite{carion2020end,sun2021rethinking,fang2021you}, semantic segmentation~\cite{xie2021segformer,zheng2021rethinking,strudel2021segmenter},
image  restoration~\cite{zamir2022restormer,liang2021swinir,chen2021pre}, etc.
Among these, the most typical work should be Vision Transformer (ViT)~\cite{dosovitskiy2020image}, which proves Transformers can perform better
than convolutional neural networks on feature encoding.
The application of Transformers in low-level vision mainly includes two categories: generation~\cite{jiang2021transgan,lee2021vitgan,zhang2022styleswin,deng2022stytr2} and restoration.
Further, the restoration tasks can also be divided into two categories: video restoration~\cite{lu2022video,liu2021fuseformer,ren2022dlformer,liu2022learning,geng2022rstt} and image restoration~\cite{chen2021pre,zamir2022restormer,wang2022uformer,guo2022image}.

Image super-resolution is an important task in image restoration. 
It needs to preserve the structural information of the input, which is a great challenge for Transformer-based model design.
IPT~\cite{chen2021pre} is a large pre-trained model based on the Transformer encoder and decoder structure and has been applied to super-resolution, denoising, and deraining.
Based on the Swin Transformer encoder~\cite{liu2021swin}, SwinIR~\cite{liang2021swinir}  performs self-attention on an $8\times8$ local window in feature extraction and achieves extremely powerful performance.
ELAN~\cite{zhang2022efficient} simplifies the architecture of SwinIR and uses self-attention computed in different window sizes to collect the correlations between long-range pixels.

Our \namesrformer{} is also based on Transformer.
Unlike the methods above that directly leverage self-attention to build models, our \namesrformer{}  mainly aims at the self-attention itself.
We intend to study how to compute self-attention in a large window to improve the performance of SR models without increasing the parameters and computational cost.

\section{Method}

\subsection{Overall Architecture}

The overall architecture of our \namesrformer{} and \nameofmethod{} is shown in \figref{fig:arch}, consisting of three parts: a pixel embedding layer $G_{P}$, a feature encoder $G_{E}$, and a high-resolution image reconstruction layer $G_{R}$. 
Following previous works~\cite{liang2021swinir,zhang2022efficient}, the pixel embedding layer $G_{P}$ is a single $3\times3$ convolution that transforms the low-resolution RGB image $I \in \mathbb{R}^{H \times W \times 3}$ to feature embeddings $F_{P} \in \mathbb{R}^{H \times W \times C}$.
$F_{P}$ will then be sent into the feature encoder $G_E$ with a hierarchical structure.
It consists of $N$ permuted self-attention groups, each of which is with $M$ permuted self-attention blocks followed by a $3\times3$ convolution. 
A $3\times3$ convolution is added at the end of the feature encoder, yielding
$F_E$. 
The summation results of $F_E$ and $F_{P}$ are fed into $G_R$ for high-resolution image reconstruction, which contains a $3\times3$ convolution and a sub-pixel convolutional layer~\cite{shi2016real} to reconstruct high-resolution images.
We compute the L1 loss between the high-resolution reconstructed image and ground-truth HR image to optimize our models.

\begin{figure*}
    \centering
    \small
    \setlength{\abovecaptionskip}{0pt}
    \includegraphics[width=0.95\linewidth]{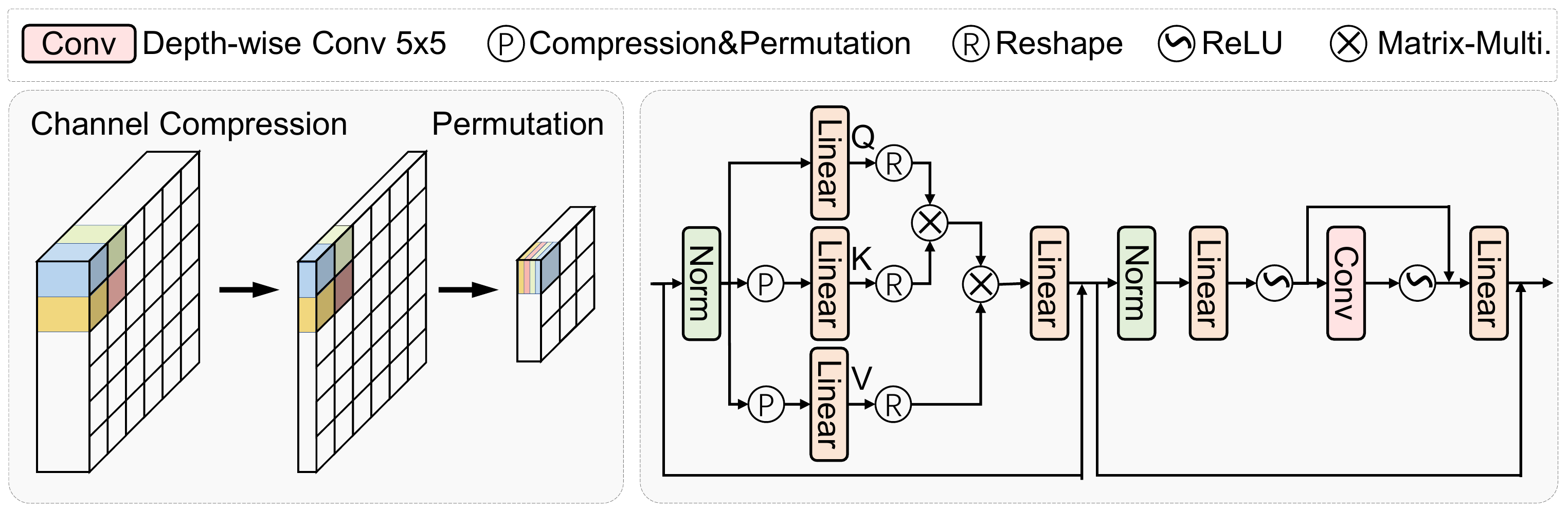}
    \caption{Comparison between (a) self-attention and (b) our proposed permuted self-attention. To avoid  spatial information loss, we propose to reduce the channel numbers and transfer the spatial information to the channel dimension.}
    \label{fig:psa_comp}
\end{figure*}

\subsection{Permuted Self-Attention Block}\label{sec:PSA}

The core of our \namesrformer{} and \nameofmethod{} is the permuted self-attention block (PAB), which consists of a permuted self-attention (PSA) layer and a convolutional feed-forward network (ConvFFN).

\myPara{Permuted self-attention.} As shown in  \figref{fig:psa_comp}(b),  given an input feature map $\mathbf{X}_{in} \in \mathbb{R}^{H \times W \times C}$ and a tokens reduction factor $r$, we first split $\mathbf{X}_{in}$ into $N$ non-overlapping square windows  $\mathbf{X} \in \mathbb{R}^{NS^2 \times C}$, where $S$ is the side length of each window.
Then, we use three linear layers $L_Q,L_K,L_V$ to get $\mathbf{Q} $, $\mathbf{K}$, and   $\mathbf{V}$:
\begin{equation}
\mathbf{Q},\mathbf{K},\mathbf{V} = L_Q(\mathbf{X}),L_K(\mathbf{X}), L_V(\mathbf{X}).
\end{equation}
Here, $\mathbf{Q}$ keeps the same channel dimension to $\mathbf{X}$ while $L_K$ and $L_V$ compress the channel dimension to $C/r^2$, yielding $\mathbf{K} \in \mathbb{R}^{NS^2 \times C/r^2}$  and   $\mathbf{V} \in \mathbb{R}^{NS^2  \times C /r^2}$. 
After that, to enable more tokens to get involved in the self-attention calculation and avoid the increase of the computational cost, we propose to permute the spatial tokens in $\mathbf{K}$ and $\mathbf{V}$ to the channel dimension, attaining permuted tokens $\mathbf{K}_{p} \in \mathbb{R}^{NS^2/r^2 \times C}$ and  $\mathbf{V}_{p} \in \mathbb{R}^{NS^2 /r^2 \times C}$. 

We use $\mathbf{Q}$ and the shrunken $\mathbf{K}_{p}$ and $\mathbf{V}_{p}$ to perform the self-attention operation.
In this way, the window size for $\mathbf{K}_{p}$ and $\mathbf{V}_{p}$
will be reduced to $\frac{S}{r}\times\frac{S}{r}$ but their channel dimension is still unchanged to guarantee the expressiveness of the attention map generated by each attention head~\cite{touvron2021going}.
The formulation of the proposed PSA can be written as follows:
\begin{equation}
\operatorname{PSA}(\mathbf{Q}, \mathbf{K}_{p}, \mathbf{V}_{p})=\operatorname{Softmax}\left(\frac{\mathbf{Q}  \mathbf{K}_{p}^{T}}{\sqrt{d_{k}}}+\mathbf{B}\right)  \mathbf{V}_{p},
\end{equation}
where $\mathbf{B}$ is an aligned relative position embedding that can be attained by interpolating the original one defined in~\cite{liu2021swin} since the window size of $\mathbf{Q}$ does not match that of $\mathbf{K}_{p}$.
$\sqrt{d_{k}}$ is a scalar as defined in~\cite{dosovitskiy2020image}.
Note that the above equation can easily be converted to the multi-head version by splitting the channels into multiple groups.
\begin{figure}[!tbp]
    \vspace{-5pt}
    \centering
    \includegraphics[width=\linewidth]{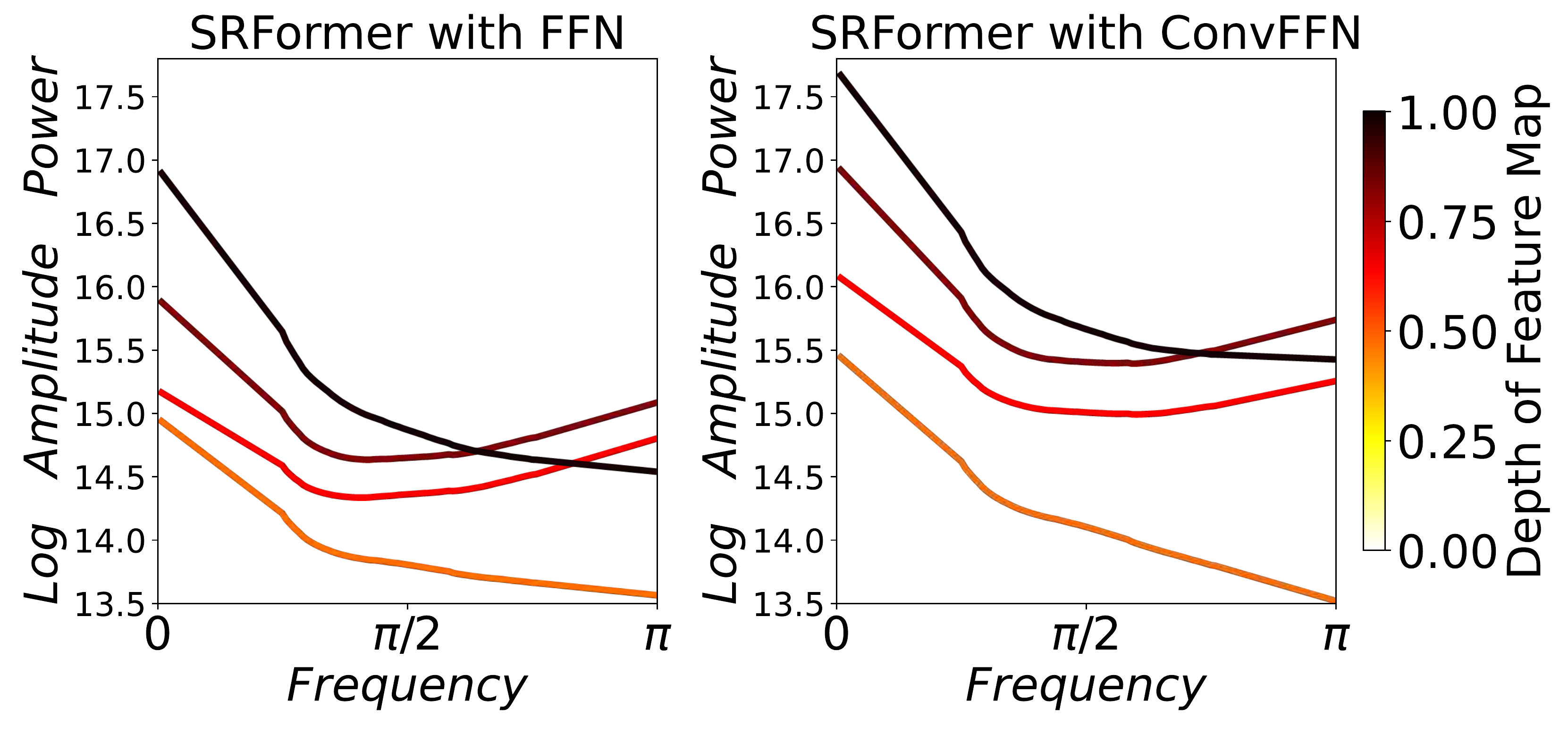}
    \vspace{-22pt}
    \caption{Power spectrum of the intermediate feature maps produced by our \namesrformer{}  with FFN and ConvFFN. Lines in darker  color correspond to features from deeper layers.}
    \label{fig:power}
    \vspace{-5pt}
\end{figure}  
Our PSA transfers the spatial information to the channel dimension.
It ensures the following two key design principles: i) We do not downsample the tokens first as done in~\cite{xie2021segformer,wang2021pyramid} but allow each token to participate in the self-attention computation independently. This enables more representative attention maps.
We will discuss more variants of our PSA in \secref{sec:variants} and show more results in our experiment section.
ii) In contrast to the original self-attention illustrated in \figref{fig:psa_comp}(a), PSA can be conducted in a large window (e.g., $24\times24$) using even fewer computations than SwinIR with $8\times8$ window while attaining better performance. 

\myPara{ConvFFN.} Previous works have demonstrated that self-attention can be viewed as a low-pass filter~\cite{parkvision,wang2022anti}.
To better restore high-frequency information, a $3\times3$ convolution is often added at the end of each group of Transformers as done in SwinIR~\cite{liang2021swinir}.
Different from SwinIR, in our PAB, we propose to add a local depthwise convolution branch
between the two linear layers of the FFN block to assist in encoding more details.
We name the new block ConvFFN.
We empirically found that such an operation increases nearly no computations but can
compensate for the loss of high-frequency information caused by self-attention shown in \figref{fig:power}. 
We simply calculate the power spectrum of the feature maps produced by our \namesrformer{}  with FFN and ConvFFN.
By comparing the two figures, we can see that ConvFFN can clearly increase high-frequency information, and hence yields better results as listed in \tabref{tab:window_size}.

\subsection{Large-Window Self-Attention Variants} \label{sec:variants}

To  provide guidance for the design of large-window self-attention and demonstrate the advantage of our PSA, here, we introduce another two large-window self-attention variants.
The quantitative comparisons and  analysis can be found in our experiment section.

\myPara{Token Reduction.} The first way to introduce large-window self-attention and avoid the increase in computational cost is to reduce the number of tokens as done in~\cite{xie2021segformer}.
Let $r$ and $S$ be a reduction factor and the window size.
Given an input $\mathbf{X}\in \mathbb{R}^{N\times S^2 \times C}$, we can adopt a depthwise convolutional function with kernel size $r \times r$ and stride $r$ to reduce the token  numbers of $\mathbf{K}$ and $\mathbf{V}$ in each window to $(\frac{S}{r})^2$, yielding
$\mathbf{Q}_r \in \mathbb{R}^{N\times S^2 \times C}$, $\mathbf{K}_r \in \mathbb{R}^{N \times S^2/r^2 \times C}$, and $\mathbf{V}_r \in \mathbb{R}^{N \times S^2/r^2 \times C}$.
$\mathbf{Q}_r$ and $\mathbf{K}_r$ are used to compute the attention scores $\mathbf{A} \in \mathbb{R}^{{N \times S^2} \times {S^2/r^2} }$.
Computing the matrix multiplication between $\mathbf{A}$ and  $\mathbf{V}_r$ yields
the output with the same number of tokens to $\mathbf{X}$.

\myPara{Token Sampling.} The second way to achieve large-window self-attention is to randomly sample $T^2~(0 \le T \le S)$ tokens from each window according to a given sampling ratio $t$ for the key $\mathbf{K}$ and value $\mathbf{V}$.
Given the input $\mathbf{X}\in \mathbb{R}^{NS^2 \times C}$, $\mathbf{Q}$ shares the same shape with $\mathbf{X}$ but the shapes of $\mathbf{K}$ and $\mathbf{V}$
are reduced to $NT^2 \times C$.
In this way, as long as $T$ is fixed, the computational cost increases linearly as the window size gets larger.
A drawback of token sampling is that randomly selecting a portion of tokens loses structural information of the scene content, which is essentially needed for image super-resolution.
We will show more numerical results in our experiment section.

\begin{figure}
    \centering
    \scriptsize
    \includegraphics[width=\linewidth]{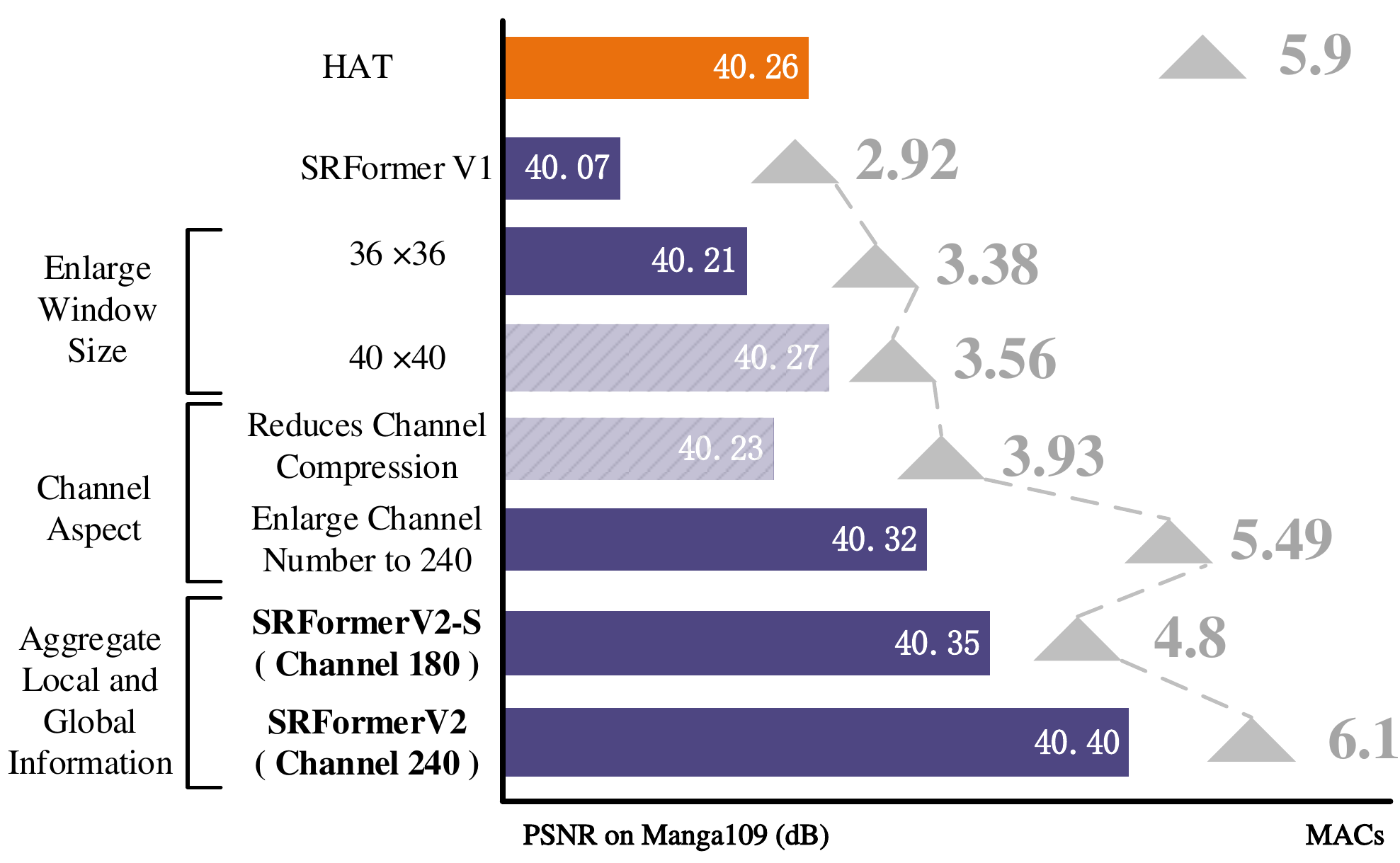}

    \caption{ The enhancement roadmap for \nameofmethod{} begins with the original \namesrformer{}. Through multifaceted improvements, we scale our SRFormer up and achieve significant performance enhancements. The backslash 
 block indicates enhancement methods can also improve performance but are not adopted in \nameofmethod{}. 
The orange square represents the previous state-of-the-art (SOTA) methods, HAT~\cite{chen2022activating}.}
    \label{fig:roadmap}
\end{figure}

\section{\nameofmethod{}: Scaling the SRFormer} 
\label{sec:srformerv2}
In this section, we further explore the potential of our SRFormer by scaling it up as shown in \figref{fig:roadmap} and present a new version, \nameofmethod{}.
We intend to figure out two things:
i) Whether PSA with a larger window can further boost the performance;
ii) How to balance the channel and spatial information, as well as the local and global information to maximize the effect of PSA.

\myPara{Scaling the window size.} The original \namesrformer{} employs permuted self-attention to perform SR reconstruction efficiently. By compressing the channel information and transferring spatial information, PSA can enjoy large-window self-attention with fewer computations, boosting the performance of previous Transformer-based models.
Leveraging the PSA, we can enjoy large window self-attention with acceptable computations.
Thus, it is natural to investigate whether larger windows could further lift the performance.

To figure it out, we conduct more experiments to further enlarge the window size, as shown in \tabref{tab:enlarge_win}.
We find that for the low-resolution test set, such as Set5 or B100, the performance no longer improves when we scale the window size up to $36\times36$.
However, we find that in high-resolution test datasets, such as Urban100 or Manga109, even when we reach the maximum level allowed by our computational resources, a window size of $40\times40$, notable improvements in performance can still be achieved.
This finding provides strong evidence for the significance of larger attention windows in image super-resolution, which is inspiring for super-resolution backbone design.
It is worth noting that, by leveraging PSA, our computations are still reasonable even with the $40\times40$ window size,  more efficient and effective than the recent state-of-the-art model HAT.
In the end, we set the window size of $36\times36$ to strike a balance between computational cost and performance.

Another advantage we discover when increasing the window size is the superiority of the structural similarity index measure (SSIM).
As shown in \tabref{tab:classical_SR}, our \namesrformer{} almost always achieves the highest SSIM values, even in cases where our Peak Signal-to-Noise Ratio (PSNR) values are similar to those of the comparison models. 
SSIM~\cite{wang2004image} is proposed to measure the differences between the structure properties of the areas. Instead of simply computing the absolute error between the images like PSNR, SSIM is proposed to take into account known properties of the human visual system.

\myPara{Scaling in the channel aspect.} Another scaling exploration is to enlarge the channel number. The original PSA compresses the channel number, so there is some sacrifice in channel information. By scaling the channel number, we hope to compensate for the loss of channel information. As shown in the channel aspect bracket of \figref{fig:roadmap}, two explorations are carried out: 
i) We enlarge the channel number to preserve more channel information. Increasing the number of channels allows the use of more information during inference, which results in significant performance gains. However, it also leads to an increase in computational load. 
By integrating the operations of scaling the channel size and the scaling window size mentioned earlier, we can achieve significant improvements compared to the original results of \namesrformer{}.
ii) Reduce the degree of channel compression. Our permuted self-attention compresses the channel dimension and transfers the spatial information to the channel dimension. By increasing the number of channels after compression, we can reduce the loss of channel information. This brings some improvements, but it is not adopted in \nameofmethod{} after taking account of the performance improvement and computing overhead.

\myPara{Aggregating Global and Local information.}
Based on PSA, we are able to facilitate a broad range of information interchange with a very small computational cost. However, as shown in \cite{kong2022residual}, the local information also plays a significant role in enhancing the performance of SR.
During the expiration, we hope to introduce the local spatial information into our large window self-attention SR network.
We rethink the macro architecture of the network and attempt to involve more local information. Following previous SR models~\cite{liang2021swinir}, the size of attention windows in ~\namesrformer{} is all the same. 
However, due to the advantage of our PSA, our attention window size is particularly large, as it is conducive to aggregating global information. In order to further improve performance, we contemplate whether we could also introduce some blocks with small attention windows to help capture local information. Therefore, prior to every two permute self-attention blocks (abbreviated as PAB shown in \figref{fig:arch} ), we insert a block with a small window size, to aggregate local spatial information before gathering global information. We found that such an approach is particularly useful for enhancing performance. 

By combining the aforementioned enhancement methods as shown in \figref{fig:roadmap}, we have upgraded \namesrformer{} to \nameofmethod{}, achieving a new \sArt. With a similar computational cost, it largely improves the previous \sArt~HAT. We name the scaling version as \nameofmethod{}. We also design a version with even less computational cost named \nameofmethod{-S}, which still achieves comparable performance with HAT.

\begin{table*}[tp]
\setlength\tabcolsep{3.5pt}
\renewcommand{\arraystretch}{1.0}
\centering
\setlength{\belowcaptionskip}{0cm}   
\caption{
 Ablation study on the window size. We report results on the original SwinIR, \namesrformer{} without ConvFFN, and our full \namesrformer{}. Note that the parameters and MACs of \namesrformer{} with $24\times24$ window are fewer than SwinIR with $8\times8$ window. Larger windows can result in better performance.
}
\begin{tabular}{lccccccccccccc}
\toprule
\multirow{2}{*}{Method} & \multirow{2}{*}{Window size} & \multirow{2}{*}{Params} & \multirow{2}{*}{MACs}  &  \multicolumn{2}{l}{  \makebox[0.10\textwidth][c]{SET5~\cite{bevilacqua2012low}} } & \multicolumn{2}{l}{\makebox[0.10\textwidth][c]{SET14~\cite{zeyde2010single}}} & \multicolumn{2}{l}{ \makebox[0.10\textwidth][c]{B100~\cite{martin2001database}}} & \multicolumn{2}{l}{\makebox[0.10\textwidth][c]{Urban100~\cite{huang2015single}}} & \multicolumn{2}{l}{\makebox[0.10\textwidth][c]{Manga109~\cite{matsui2017sketch}}} \\ 
\cmidrule(lr){5-6}\cmidrule(lr){7-8}\cmidrule(lr){9-10}\cmidrule(lr){11-12} \cmidrule(lr){13-14}
& & & &\makecell{PSNR}  &\makecell{SSIM} &\makecell{PSNR}  &\makecell{SSIM}&\makecell{PSNR}  &\makecell{SSIM}  &\makecell{PSNR}  &\makecell{SSIM}&\makecell{PSNR}  &\makecell{SSIM}     \\ \midrule
\multirow{3}{*}{$ \rm \textbf{SwinIR}$~\cite{liang2021swinir}}& $8\times8$ &11.75M&2868G&38.24&0.9615&33.94&0.9212&32.39&0.9023&33.09&0.9373&39.34&0.9784\\
& $12\times12$ &11.82M&3107G&38.30&0.9617&34.04&0.9220&32.42&0.9026&33.28&0.9381&39.44&0.9788\\
& $16\times16$ &11.91M&3441G&38.32&0.9618&34.00&0.9212&32.44&0.9030&33.40&0.9394&39.53&0.9791
\\ \midrule
\multirow{3}{*}{\makecell{$\rm \textbf{\namesrformer{}}$ w/o $\textbf{ConvFFN}$}} 
& $12\times12$  &9.97M&2381G&38.23&0.9615&34.00&0.9216&32.37&0.9023&32.99&0.9367&39.30&0.9786 \\
&  $16\times16$ &9.99M&2465G&38.25&0.9616&33.98&0.9209&32.38&0.9022&33.09&0.9371&39.42&0.9789  \\
&  $24\times24$ &10.06M&2703G&38.30&0.9618&34.08&0.9225&32.43&0.9030&33.38&0.9397&39.44&0.9786    \\ \midrule
\multirow{3}{*}{$\rm \textbf{\namesrformer{}}$}
& $12\times12$  &10.31M&2419G&38.22&0.9614&34.08&0.9220&32.38&0.9025&33.08&0.9372&39.13&0.9780 \\
&  $16\times16$ &10.33M&2502G&38.31&0.9617&34.10&0.9217&32.43&0.9026&33.26&0.9385&39.36&0.9785 \\
&  $24\times24$  &10.40M&2741G&38.33&0.9618&34.13&0.9228&32.44&0.9030&33.51&0.9405&39.49&0.9788    \\
 \bottomrule 
\end{tabular}

\label{tab:window_size}
\end{table*}

\begin{table}[tp]
\centering
\setlength{\belowcaptionskip}{0cm}   
\setlength\tabcolsep{7.5pt}
\renewcommand{\arraystretch}{1.0}
\caption{
 Ablation study on ConvFFN for $\times$2 SR. From the results on Urban100 and Manga109, we can see that using $5\times5$ depthwise convolution yields the best results. This indicates that local details are also essential for Transformer-based models.
}
\vspace{-5pt}
\begin{tabular}{cccccccccccccc}
\toprule
 \multirow{2}{*}{ConvFFN}  &  \multicolumn{2}{l}{\makebox[0.10\textwidth][c]{Urban100~\cite{huang2015single}}} & \multicolumn{2}{l}{\makebox[0.10\textwidth][c]{Manga109~\cite{matsui2017sketch}}} \\ 
\cmidrule(lr){2-3} \cmidrule(lr){4-5}
& \makecell{PSNR}  &\makecell{SSIM} &\makecell{PSNR}  &\makecell{SSIM}  \\ \midrule
  w/o Depth-wise Conv &33.38&0.9397&39.44&0.9786\\
 $3\times3$ Depth-wise Conv &33.42&0.9398&39.34&0.9787\\
$5\times5$ Depth-wise Conv &\highlight{33.51}&\highlight{0.9405}&\highlight{39.49}&\highlight{0.9788}\\
 \bottomrule 
\end{tabular}
\vspace{-8pt}
\label{tab:depth-wise}
\end{table}

\begin{table}[tp]
\setlength\tabcolsep{4.5pt}
\renewcommand{\arraystretch}{1.0}
\small
\centering
\setlength{\belowcaptionskip}{0cm}   
\caption{
 $\times2$ SR performance comparison among SwinIR~\cite{liang2021swinir}, our proposed PSA, and the two variants on Urban100~\cite{huang2015single}. The results reported here are based on the best model trained on DIV2K for 200k iterations.  $r$ and $S$ are the reduction factor and the window size. For token sampling, $r=S/T$. PSA performs better than the other two variants.
}
\vspace{-5pt}
\begin{tabular}{lcccccc}
\toprule
Method  & Params & MACs & $S$ & $r$ & PSNR & SSIM  \\ 
\midrule
 SwinIR~\cite{liang2021swinir}& 11.75M&2868G& 8 &	- &33.09& 	0.9373
\\ \midrule
Token Reduction & 11,78M & 2471G & 16 & 2 &	33.09  &	0.9372     \\
Token Reduction & 11.85M & 2709G & 24 & 2 &	33.24  &	0.9387     \\  \midrule
Token Sampling &11.91M&2465G& 16 & 2 & 32.38&0.9312   \\
Token Sampling &12.18M&2703G& 24 & 2 & 32.34&0.9305   \\  \midrule
PSA & 9.99M &	2465G	& 16 & 2 & 33.09 &	0.9371	\\
PSA & 9.67M&	2390G	& 24 & 3 &33.09 &	0.9376	\\
PSA & 10.06M&	2703G	& 24 & 2 &33.38 &	0.9397	\\ \bottomrule %
\end{tabular}

\vspace{-5pt}
\label{tab:large_window_variant}
\end{table}

\begin{table}[htp!]
\setlength\tabcolsep{3.0pt}
\renewcommand{\arraystretch}{1}
\small
\centering
\setlength{\belowcaptionskip}{0cm}   
\caption{
We further scale the window size of \namesrformer{} to explore our \nameofmethod{}. 
We report PSNR on Set5, B100, Urban100, and Manga109. All models are trained on the DF2K (DIV2K~\cite{lim2017enhanced} + Flickr2K~\cite{timofte2017ntire}) training set for 500k training iterators.
}
\vspace{-5pt}
\begin{tabular}{cccccc}
\toprule
Window Size &MACs & Set5 & B100& Urban100 &Manga109 \\ \midrule
$24\times24$  &2741G &38.53&32.59 &34.09 &40.07 \\ 
$36\times36$ & 3377G &38.58 &32.61&34.39&40.21 \\ 
$40\times40$& 3562G&38.58 &32.61& 34.52 & 40.27 \\ 
\bottomrule
\end{tabular}
\label{tab:enlarge_win}
\vspace{-10pt}
\end{table}

\section{Experiments}
In this section, we conduct experiments on both the classical, lightweight, and real-world image SR tasks, compare our \namesrformer{} and \nameofmethod{}  with existing state-of-the-art methods, and do ablation analysis of the proposed method.

\subsection{Experimental Setup}
\myPara{Datasets and Evaluation.}
The choice of training datasets remains the same as the comparison models. In classical image SR, we use DIV2K~\cite{lim2017enhanced} and DF2K (DIV2K~\cite{lim2017enhanced} + Flickr2K~\cite{timofte2017ntire}) to train two versions of \namesrformer{}. In the training of \nameofmethod{}, we also use  DF2K as the training dataset.
For lightweight image SR, we use DIV2K~\cite{lim2017enhanced} to train our \namesrformer{}-light.
For real-world SR, we use DF2K and OST~\cite{wang2018recovering}. 
For testing, we mainly evaluate our method on five benchmark datasets, including Set5~\cite{bevilacqua2012low}, Set14~\cite{zeyde2010single}, BSD100~\cite{martin2001database}, Urban100~\cite{huang2015single}, and Manga109~\cite{matsui2017sketch}. Self-ensemble strategy is introduced in \nameofmethod{} to further improve performance, marked as \nameofmethod{+}.
The experimental results are evaluated in terms of PSNR and SSIM values, which are calculated on the Y channel from the YCbCr space.

\myPara{Implementation Details.} In the classical image SR task, we set the PAB group number, PAB number, channel number, and attention head number to 6, 6, 180, and 6, respectively.
%
%
When training \namesrformer{} on DIV2K~\cite{lim2017enhanced}, the patch size, window size $S$, and reduction factor $r$ are set to $48\times48$, 24, and 2, respectively. 
When training  \namesrformer{} on the DF2K~\cite{lim2017enhanced,timofte2017ntire}, they are $64\times64$, 22, and 2, respectively.
For the training of \nameofmethod{},   
we also follow the same training settings.
For the lightweight image SR task, we set the PAB group number, PAB number, channel number, windows size $S$, reduction factor $r$, and attention head number to 4, 6, 60, 16,  2, and 6, respectively.
The training patch size we use is $64\times 64$. 
%
We randomly rotate images by $90^{\circ}$, $180^{\circ}$, or $270^{\circ}$ and randomly flip images horizontally for data augmentation.
We adopt the Adam~\cite{kingma2014adam} optimizer with $\beta_1 = 0.9$ and  $\beta_2 = 0.99$ to train the model for 500k iterations.  The initial learning rate is set as $2\times10^{-4}$ and is reduced by half at the $\rm \left\{{250k,400k,450k,475k}\right\}$-th iterations.

\subsection{Ablation Study}

\myPara{Impact of window size in PSA.} Permuted self-attention provides an efficient and effective way to enlarge window size.
To investigate the impact of different window sizes on model performance, we conduct three groups of experiments and report the results in Table \ref{tab:window_size}.
The first group is the vanilla SwinIR~\cite{liang2021swinir} with $8 \times 8$, $12\times12$, and $16\times16$ window sizes.
In the second group, we do not use the ConvFFN but only the PSA in our \namesrformer{} and set the window size to $12\times12$, $16\times16$, and $24\times24$, respectively, to observe the performance difference.
In the third group, we use our full \namesrformer{} with $12\times12$, $16\times16$, and $24\times24$ as window size to explore the performance change.
The results show that a larger window size yields better performance improvement for all three groups of experiments.
In addition, the parameters and MACs of our \namesrformer{} with $24\times24$ window are even fewer than the original SwinIR with $8\times8$ window.
To balance the performance and MACs, we set window size as $24\times24$  in \namesrformer{} and $16\times16$  in \namesrformer{-light}.

\myPara{Impact of kernel size of ConvFFN.}
We introduce ConvFFN in \secref{sec:PSA}, which aims to encode more local information without increasing too many computations.
In order to explore which kernel size can bring the best performance improvement, 
we attempt to use $3\times3$ depth-wise convolution and $5\times5$ depth-wise convolution and report the results in Table \ref{tab:depth-wise}. 
Given that the depth-wise convolution has little effect on the number of parameters and MACs, we do not list them in the table.
Obviously, $5\times5$ depth-wise convolution leads to the best results.
Thus, we use $5\times5$ depth-wise convolution in our ConvFFN.

\myPara{Large-window self-attention variants.} In \secref{sec:variants}, we introduce another two large-window self-attention variants.
We summarize the results in Table~\ref{tab:large_window_variant}.
Though token reduction can slightly improve SwinIR when using a large window,
the number of parameters does not decrease and the performance gain is lower than ours. We argue that it is because directly applying downsampling operations to the key and value results in spatial information loss.
For token sampling, the performance is even worse than the original SwinIR.
We believe the reason is that dropping out some tokens severely breaks the image content structure.

\begin{table*}[!htp]
\setlength\tabcolsep{6.9pt}
\renewcommand{\arraystretch}{1}
\footnotesize
\centering
\setlength{\belowcaptionskip}{0cm}   
\caption{
Quantitative comparison of our \namesrformer{} with recent \sArt~\textbf{ {classical image SR}} methods on five benchmark datasets.
The best performance is highlighted. 
}
\begin{tabular}{crccccccccccc}
\toprule
& \multirow{2}{*}{Method}  & \multirow{2}{*}{\makecell{Training \\ Dataset}} &  \multicolumn{2}{c}{  SET5~\cite{bevilacqua2012low} } & \multicolumn{2}{c}{SET14~\cite{zeyde2010single}} & \multicolumn{2}{c}{ B100~\cite{martin2001database}} & \multicolumn{2}{c}{Urban100~\cite{huang2015single}} & \multicolumn{2}{c}{Manga109~\cite{matsui2017sketch}} \\ 
\cmidrule(lr){4-5}\cmidrule(lr){6-7}\cmidrule(lr){8-9}\cmidrule(lr){10-11} \cmidrule(lr){12-13} 
    & & &\makecell{PSNR}  &\makecell{SSIM} &\makecell{PSNR}  &\makecell{SSIM}&\makecell{PSNR}  &\makecell{SSIM}  &\makecell{PSNR}  &\makecell{SSIM}&\makecell{PSNR}  &\makecell{SSIM}     \\ \midrule
\multirow{11}{*}{\rotatebox{90}{$\times 2$ SR}} 
& $\rm \textbf {EDSR}$~\cite{lim2017enhanced}& DIV2K  &38.11&0.9602&33.92&0.9195&32.32&0.9013&32.93&0.9351&39.10&0.9773   \\
& $\rm \textbf{RCAN}$~\cite{zhang2018image} & DIV2K  & 38.27&0.9614&34.12&0.9216&32.41&0.9027&33.34&0.9384&39.44&0.9786   \\
& $\rm \textbf{SAN}$~\cite{dai2019second} & DIV2K &38.31&0.9620&34.07&0.9213&32.42&0.9028&33.10&0.9370&39.32&0.9792    \\
& $\rm \textbf{IGNN}$~\cite{zhou2020cross}& DIV2K &38.24&0.9613&34.07&0.9217&32.41&0.9025&33.23&0.9383&39.35&0.9786\\
& $\rm \textbf{HAN}$~\cite{niu2020single}& DIV2K &38.27&0.9614&34.16&0.9217&32.41&0.9027&33.35&0.9385&39.46&0.9785\\
& $\rm \textbf{NLSA}$~\cite{mei2021image}& DIV2K  &38.34&0.9618&34.08&0.9231&32.43&0.9027&33.42&0.9394&39.59&0.9789   \\
& $\rm \textbf{SwinIR}$~\cite{liang2021swinir}& DIV2K  &38.35&0.9620&34.14&0.9227&32.44&0.9030&33.40&0.9393&39.60&0.9792 \\
& $\rm \textbf{ELAN}$~\cite{zhang2022efficient}& DIV2K &38.36&0.9620&34.20&0.9228&32.45&0.9030&33.44&0.9391&39.62&\highlight{0.9793} \\ 
& \textbf{\namesrformer{ (ours)}}& DIV2K & \highlight{38.45}&\highlight{0.9622}&\highlight{34.21}&\highlight{0.9236}&\highlight{32.51}&\highlight{0.9038}&\highlight{33.86}&\highlight{0.9426}&\highlight{39.69}&0.9786 \\ 
 \cmidrule(lr){2-13}
& $\rm \textbf{IPT}$~\cite{chen2021pre}& ImageNet  &38.37&-&34.43&-&32.48&-&33.76&-&-&-\\
& $\rm \textbf{SwinIR}$~\cite{liang2021swinir}& DF2K &38.42&0.9623&34.46&0.9250&32.53&0.9041&33.81&0.9427&39.92&0.9797\\
& $\rm \textbf{EDT}$~\cite{li2021efficient}& DF2K &38.45&0.9624&{34.57}&{0.9258}&32.52&0.9041&33.80&0.9425&39.93&0.9800\\
& $\rm \textbf{HAT}$~\cite{li2021efficient}& DF2K 
&{38.63}&0.9630	&34.86&0.9274	&32.62&0.9053	&34.45&0.9466	&40.26&0.9809\\
& \textbf{\namesrformer{ (ours)}} & DF2K & {38.51}&{0.9627}&34.44&0.9253&{32.57}&{0.9046}&{34.09}&{0.9449}&{40.07}&{0.9802}  \\
& \textbf{\nameofmethod{ (ours)}} & DF2K &  {38.63} & {0.9632} & 	{34.90} & {0.9276}	&{32.64} & {0.9056} &	{34.59}& {0.9477}	& {40.40} & {0.9811} \\
& \textbf{\nameofmethod{+ (ours)}} & DF2K & 
\highlight{38.68} & \highlight{0.9633} & \highlight{34.93} & \highlight{0.9278} & \highlight{32.67} &  \highlight{0.9059}  & \highlight{34.77} & \highlight{0.9485} & \highlight{40.51} & \highlight{0.9813}\\
\midrule
\multirow{11}{*}{\rotatebox{90}{$\times 3$ SR}} 
& $ \rm \textbf {EDSR}$~\cite{lim2017enhanced}& DIV2K  &34.65&0.9280&30.52&0.8462&29.25&0.8093&28.80&0.8653&34.17&0.9476  \\
& $\rm \textbf{RCAN}$~\cite{zhang2018image} & DIV2K &34.74&0.9299&30.65&0.8482&29.32&0.8111&29.09&0.8702&34.44&0.9499 \\
& $\rm \textbf{SAN}$~\cite{dai2019second} & DIV2K  &34.75&0.9300&30.59&0.8476&29.33&0.8112&28.93&0.8671&34.30&0.9494   \\
& $\rm \textbf{IGNN}$~\cite{zhou2020cross}& DIV2K  &34.72&0.9298&30.66&0.8484&29.31&0.8105&29.03&0.8696&34.39&0.9496\\
& $\rm \textbf{HAN}$~\cite{niu2020single}& DIV2K &34.75&0.9299&30.67&0.8483&29.32&0.8110&29.10&0.8705&34.48&0.9500\\
& $\rm \textbf{NLSA}$~\cite{mei2021image}& DIV2K  &34.85&0.9306&30.70&0.8485&29.34&0.8117&29.25&0.8726&34.57&0.9508   \\
& $\rm \textbf{SwinIR}$~\cite{liang2021swinir}& DIV2K &34.89&0.9312&30.77&0.8503&29.37&0.8124&29.29&0.8744&34.74&0.9518 \\
& $\rm \textbf{ELAN}$~\cite{zhang2022efficient}& DIV2K &34.90&0.9313&30.80&0.8504&29.38&0.8124&29.32&0.8745&34.73&0.9517 \\ 
& \textbf{\namesrformer{ (ours)}}& DIV2K &  \highlight{34.94}&\highlight{0.9318}&\highlight{30.81}&\highlight{0.8518}&\highlight{29.41}&\highlight{0.8142}&\highlight{29.52}&\highlight{0.8786}&\highlight{34.78}&\highlight{0.9524} \\  \cmidrule(lr){2-13}
& $\rm \textbf{IPT}$~\cite{chen2021pre}& ImageNet &34.81&-&30.85&-&29.38&-&29.49&-&-&-\\
& $\rm \textbf{SwinIR}$~\cite{liang2021swinir}& DF2K  &34.97&0.9318&30.93&0.8534&29.46&0.8145&29.75&0.8826&35.12&0.9537\\
& $\rm \textbf{EDT}$~\cite{li2021efficient}& DF2K  &34.97&0.9316&30.89&0.8527&29.44&0.8142&29.72&0.8814&35.13&0.9534\\
& $\rm \textbf{HAT}$~\cite{li2021efficient}& DF2K 
&35.07 &0.9329 &31.08 &0.8555 &29.54 &0.8167& 30.23& 0.8896 &35.53& 0.9552\\
& \textbf{\namesrformer{ (ours)}} & DF2K  &  35.02&0.9323&30.94&0.8540&29.48&0.8156&30.04&0.8865&35.26&0.9543\\
& \textbf{\nameofmethod{ (ours)}} & DF2K  &   35.05 &  {0.9331} &  {31.10} & {0.8562} & {29.55} & {0.8176} & {30.46} & {0.8924} & {35.59} & {0.9557} \\

& \textbf{\nameofmethod{+ (ours)}} & DF2K  &   \highlight{35.14} &  \highlight{0.9335} &  \highlight{31.17} &  \highlight{0.8569} &  \highlight{29.59}  &  \highlight{0.8181} &  \highlight{30.62} &  \highlight{0.8941} &  \highlight{35.73} &  \highlight{0.9563} \\
\midrule
\multirow{11}{*}{\rotatebox{90}{$\times 4$ SR}} 
& $ \rm \textbf {EDSR}$~\cite{lim2017enhanced}& DIV2K  &32.46&0.8968&28.80&0.7876&27.71&0.7420&26.64&0.8033&31.02&0.9148 \\
& $\rm \textbf{RCAN}$~\cite{zhang2018image}& DIV2K  &32.63&0.9002&28.87&0.7889&27.77&0.7436&26.82&0.8087&31.22&0.9173 \\
& $\rm \textbf{SAN}$~\cite{dai2019second} & DIV2K  &32.64&0.9003&28.92&0.7888&27.78&0.7436&26.79&0.8068&31.18&0.9169   \\
& $\rm \textbf{IGNN}$~\cite{zhou2020cross}& DIV2K  &32.57&0.8998&28.85&0.7891&27.77&0.7434&26.84&0.8090&31.28&0.9182\\
& $\rm \textbf{HAN}$~\cite{niu2020single}& DIV2K  &32.64&0.9002&28.90&0.7890&27.80&0.7442&26.85&0.8094&31.42&0.9177\\
& $\rm \textbf{NLSA}$~\cite{mei2021image}& DIV2K &32.59&0.9000&28.87&0.7891&27.78&0.7444&26.96&0.8109&31.27&0.9184   \\
& $\rm \textbf{SwinIR}$~\cite{liang2021swinir}& DIV2K &32.72&0.9021&28.94&0.7914&27.83&0.7459&27.07&0.8164&31.67&0.9226 \\
& $\rm \textbf{ELAN}$~\cite{zhang2022efficient}& DIV2K &32.75&0.9022&28.96&0.7914&27.83&0.7459&27.13&0.8167&31.68&0.9226 \\ 
& \textbf{\namesrformer{ (ours)}}& DIV2K & \highlight{32.81}&\highlight{0.9029}&\highlight{29.01}&\highlight{0.7919}&\highlight{27.85}&\highlight{0.7472}&\highlight{27.20}&\highlight{0.8189}&\highlight{31.75}&\highlight{0.9237}\\  \cmidrule(lr){2-13}
& $\rm \textbf{IPT}$~\cite{chen2021pre}& ImageNet &32.64&-&29.01&-&27.82&-&27.26&-&-&-\\
& $\rm \textbf{SwinIR}$~\cite{liang2021swinir}& DF2K &32.92&{0.9044}&{29.09}&0.7950&27.92&0.7489&27.45&0.8254&32.03&0.9260\\
& $\rm \textbf{EDT}$~\cite{li2021efficient}& DF2K &32.82&0.9031&{29.09}&0.7939&27.91&0.7483&27.46&0.8246&32.03&0.9254\\
& $\rm \textbf{HAT}$~\cite{li2021efficient}& DF2K 
&33.04&0.9056	&{29.23}&0.7973	&{28.00}&0.7517	&27.97&0.8368&	32.48&0.9292 \\
&\textbf{\namesrformer{ (ours)}}& DF2K & {32.93}&0.9041&29.08&{0.7953}&{27.94}&{0.7502}&{27.68}&{0.8311}&{32.21}&{0.9271}\\
& \textbf{\nameofmethod{ (ours)}} & DF2K & 33.06 & {0.9066} & 29.21 & {0.7978} & 27.98 & {0.7522} & {28.04} & {0.8391} & {32.52} & {0.9300} \\
& \textbf{\nameofmethod{+ (ours)}} & DF2K & \highlight{33.16} &  \highlight{0.9069}  & \highlight{29.28}&  \highlight{0.7991}  & \highlight{28.03}  &  \highlight{0.7530} & \highlight{28.17} & \highlight{0.8414} & \highlight{32.75} & \highlight{0.9313}  \\
\bottomrule 
\end{tabular}
\label{tab:classical_SR}
\end{table*}


\myPara{Further scaling window size in \nameofmethod{}.} Here, we further explore the performance of \namesrformer{} with larger attention windows in \secref{sec:srformerv2}. Building upon a baseline of $24 \times 24$, we implement attention mechanisms for window sizes of $36 \times 36$ and $40 \times 40$, respectively. As shown in \tabref{tab:enlarge_win}, for the largest window size we could achieve ($40 \times 40$), we still observe performance improvement. For low-resolution images,  increasing the window size up to 36 results in similar performance. However, when further increasing the window size to $40 \times 40$, we observe no performance gain due to the small input size.

\newcommand{\addImg}[1]{\includegraphics[width=0.315\textwidth,height=.151\textwidth]{\name #1}}
\newcommand{\addImgS}[1]{\includegraphics[width=0.13\textwidth]{\name #1}}
\newcommand{\addSubFigs}[2]{
	\renewcommand{\name}{figures/#1/}
 			\begin{tabular}{cc}
	\begin{adjustbox}{valign=t}
		\begin{tabular}{c}
			\addImg{rectangle.png} \\
			#2
		\end{tabular}
	\end{adjustbox}
	\begin{adjustbox}{valign=t}
		\begin{tabular}{ccccc}
			\addImgS{HR}&
			\addImgS{Bicuble}&
			\addImgS{EDSR}&
			\addImgS{RCAN}&
			\addImgS{NLSA}
			\\
			HR&
			Bicubic&
			EDSR~\cite{lim2017enhanced}&
			RCAN~\cite{zhang2018image}&
			NLSA~\cite{zhou2020cross}
			\\
			\addImgS{ipt}&
			\addImgS{DF2K_s64w8_SwinIR} &
   			\addImgS{HAT_SRx4} &
			\addImgS{SRformer} &
                \addImgS{SRformerv2}
			\\ 
			IPT~\cite{chen2021pre}&
			SwinIR~\cite{liang2021swinir} & 
   			HAT~\cite{chen2022activating} & 
			\textbf{\namesrformer{}} (ours) &
                \textbf{\nameofmethod{}} (ours) 
			\\
		\end{tabular}
	\end{adjustbox}
 \end{tabular}
}

\begin{figure*}[!tp]
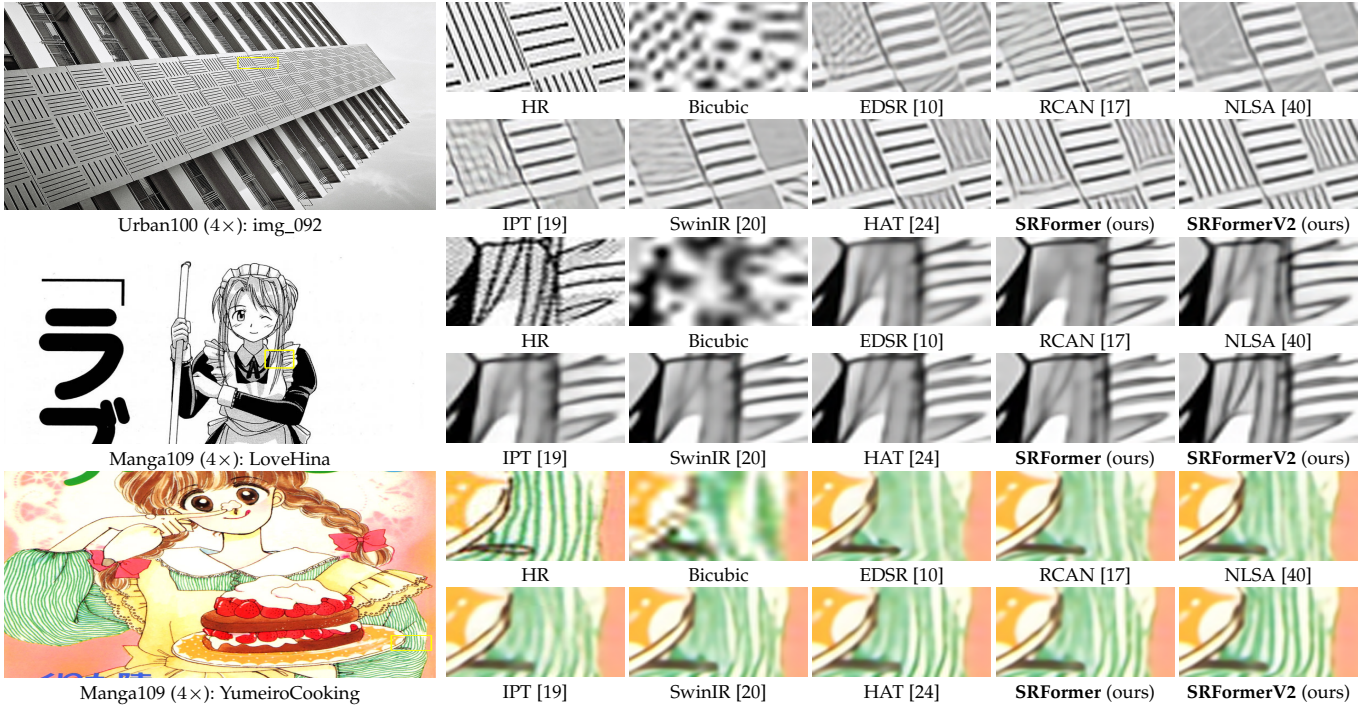

	\tablestyle{1pt}{1}
  \scriptsize{
	\newcommand{\name}{figures/classicSR/urban73/}
	\addSubFigs{classicSR/urban92}{Urban100 ($4\times$): img\_092} \\
	\addSubFigs{classicSR/manga45}{Manga109 ($4\times$): LoveHina} \\
        \addSubFigs{classicSR/manga108}{Manga109 ($4\times$): YumeiroCooking}
 }
\caption{Qualitative comparison with recent state-of-the-art  \textbf{classical image SR}  methods on the $\times 4$ SR task. }
\label{fig:sr_visual}
\end{figure*}

\begin{table*}[h!]
\setlength\tabcolsep{4.5pt}
\renewcommand{\arraystretch}{1}
\footnotesize
\centering
\caption{
Quantitative comparison of our \namesrformer{-light} with recent \sArt~\textbf{ lightweight image SR} methods on five benchmark datasets.
The best performance among all the models is highlighted.
}
\begin{tabular}{crccccccccccccc}
\toprule
& \multirow{2}{*}{Method}  & \multirow{2}{*}{\makecell{Training \\ Dataset}} & \multirow{2}{*}{Params} & \multirow{2}{*}{MACs}  &  \multicolumn{2}{c}{  SET5~\cite{bevilacqua2012low} } & \multicolumn{2}{c}{SET14~\cite{zeyde2010single}} & \multicolumn{2}{c}{ B100~\cite{martin2001database}} & \multicolumn{2}{c}{Urban100~\cite{huang2015single}} & \multicolumn{2}{c}{Manga109~\cite{matsui2017sketch}} \\ 
\cmidrule(lr){6-7}\cmidrule(lr){8-9}\cmidrule(lr){10-11} \cmidrule(lr){12-13} \cmidrule(lr){14-15}
 & &   & & &\makecell{PSNR}  &\makecell{SSIM} &\makecell{PSNR}  &\makecell{SSIM}&\makecell{PSNR}  &\makecell{SSIM}  &\makecell{PSNR}  &\makecell{SSIM}&\makecell{PSNR}  &\makecell{SSIM}     \\ \midrule
\multirow{9}{*}{\rotatebox{90}{$\times 2$ SR}} 
& $\rm \textbf{EDSR-baseline}$~\cite{lim2017enhanced} & DIV2K & 1370K &316G &37.99&0.9604&33.57&0.9175&32.16&0.8994&31.98&0.9272&38.54&0.9769 \\
& $\rm \textbf{IMDN}$~\cite{hui2019lightweight}& DIV2K & 694K & 158.8G &38.00&0.9605&33.63&0.9177&32.19&0.8996&32.17&0.9283&38.88&0.9774\\
& $\rm \textbf{LAPAR-A}$~\cite{li2020lapar}& DF2K & 548K  & 171G &38.01&0.9605&33.62&0.9183&32.19&0.8999&32.10&0.9283&38.67&0.9772\\
& $\rm \textbf{LatticeNet}$~\cite{luo2020latticenet}& DIV2K & 756K & 169.5G &38.15&0.9610&33.78&0.9193&32.25&0.9005&32.43&0.9302&-&- \\
& $\rm \textbf{ESRT}$~\cite{lu2021efficient}& DIV2K & 751K & - &38.03&0.9600&33.75&0.9184&32.25&0.9001&32.58&0.9318&39.12&0.9774\\
& $\rm \textbf{SwinIR-light}$~\cite{liang2021swinir}& DIV2K &910K & 244G  &38.14&0.9611&33.86&0.9206&32.31&0.9012&32.76&0.9340&39.12&0.9783 \\
& $\rm \textbf{ELAN}$~\cite{zhang2022efficient}& DIV2K &621K &203G &38.17&0.9611&\highlight{33.94}&0.9207&32.30&0.9012&32.76&0.9340&39.11&0.9782 \\ 
& \textbf{\namesrformer{-light}} & DIV2K & 853K & 236G & \highlight{38.23}&\highlight{0.9613}&\highlight{33.94}&\highlight{0.9209}&\highlight{32.36}&\highlight{0.9019}&\highlight{32.91}&\highlight{0.9353}&\highlight{39.28}&\highlight{0.9785} \\ \midrule
\multirow{9}{*}{\rotatebox{90}{$\times 3$ SR}} 
& $\rm \textbf{EDSR-baseline}$~\cite{lim2017enhanced} & DIV2K & 1555K & 160G&34.37&0.9270&30.28&0.8417&29.09&0.8052&28.15&0.8527&33.45&0.9439 \\
& $\rm \textbf{IMDN}$~\cite{hui2019lightweight}& DIV2K & 703K & 71.5G &34.36&0.9270&30.32&0.8417&29.09&0.8046&28.17&0.8519&33.61&0.9445\\
& $\rm \textbf{LAPAR-A}$~\cite{li2020lapar}& DF2K &594K &114G &34.36&0.9267&30.34&0.8421&29.11&0.8054&28.15&0.8523&33.51&0.9441\\
& $\rm \textbf{LatticeNet}$~\cite{luo2020latticenet}& DIV2K & 765K & 76.3G  &34.53&0.9281&30.39&0.8424&29.15&0.8059&28.33&0.8538&-&-  \\
& $\rm \textbf{ESRT}$~\cite{lu2021efficient}& DIV2K & 751K & - &34.42&0.9268&30.43&0.8433&29.15&0.8063&28.46&0.8574&33.95&0.9455\\
& $\rm \textbf{SwinIR-light}$~\cite{liang2021swinir}& DIV2K &918K & 111G &34.62&0.9289&30.54&0.8463&29.20&0.8082&28.66&0.8624&33.98&0.9478 \\
& $\rm \textbf{ELAN}$~\cite{zhang2022efficient}& DIV2K &629K &90.1G &34.61&0.9288&30.55&0.8463&29.21&0.8081&28.69&0.8624&34.00&0.9478 \\ 
& \textbf{\namesrformer{-light}} & DIV2K & 861K & 105G & \highlight{34.67}&\highlight{0.9296}&\highlight{30.57}&\highlight{0.8469}&\highlight{29.26}&\highlight{0.8099}&\highlight{28.81}&\highlight{0.8655}&\highlight{34.19}&\highlight{0.9489} \\ \midrule
\multirow{9}{*}{\rotatebox{90}{$\times 4$ SR}} 
& $\rm \textbf{EDSR-baseline}$~\cite{lim2017enhanced} & DIV2K & 1518K&114G &32.09&0.8938&28.58&0.7813&27.57&0.7357&26.04&0.7849&30.35&0.9067 \\
& $\rm \textbf{IMDN}$~\cite{hui2019lightweight}& DIV2K & 715K & 40.9G &32.21&0.8948&28.58&0.7811&27.56&0.7353&26.04&0.7838&30.45&0.9075\\
& $\rm \textbf{LAPAR-A}$~\cite{li2020lapar}& DF2K &659K & 94G &32.15&0.8944&28.61&0.7818&27.61&0.7366&26.14&0.7871&30.42&0.9074\\
& $\rm \textbf{LatticeNet}$~\cite{luo2020latticenet}& DIV2K & 777K & 43.6G &32.30&0.8962&28.68&0.7830&27.62&0.7367&26.25&0.7873&-&-  \\
& $\rm \textbf{ESRT}$~\cite{lu2021efficient}& DIV2K & 751K & - &32.19&0.8947&28.69&0.7833&27.69&0.7379&26.39&0.7962&30.75&0.9100\\
& $\rm \textbf{SwinIR-light}$~\cite{liang2021swinir}& DIV2K &930K & 63.6G &32.44&0.8976&28.77&0.7858&27.69&0.7406&26.47&0.7980&30.92&0.9151 \\
& $\rm \textbf{ELAN}$~\cite{zhang2022efficient}& DIV2K &640K &54.1G &32.43&0.8975&28.78&0.7858&27.69&0.7406&26.54&0.7982&30.92&0.9150 \\ 
& \textbf{\namesrformer{-light}}& DIV2K & 873K & 62.8G & \highlight{32.51}&\highlight{0.8988}&\highlight{28.82}&\highlight{0.7872}&\highlight{27.73}&\highlight{0.7422}&\highlight{26.67}&\highlight{0.8032}&\highlight{31.17}&\highlight{0.9165}\\ \bottomrule 
\end{tabular}
\label{tab:light_weight_SR}
\end{table*}


\subsection{Classical Image Super-Resolution}

To evaluate the performance of \namesrformer{} and \nameofmethod{} on classical super-resolution task, we compare them with a series of \sArt~CNN-based and Transformer-based SR methods: RCAN~\cite{zhang2018image}, RDN~\cite{zhang2018residual}, SAN~\cite{dai2019second}, IGNN~\cite{zhou2020cross}, HAN~\cite{niu2020single}, NLSA~\cite{mei2021image}, IPT~\cite{chen2021pre}, SwinIR~\cite{liang2021swinir}, EDT~\cite{li2021efficient}, ELAN~\cite{zhang2022efficient}, and HAT~\cite{chen2022activating}.

\myPara{Quantitative comparison.} The quantitative comparison of the methods for classical image SR is shown in Table \ref{tab:classical_SR}. For a fair comparison, the number of parameters and MACs of \namesrformer{}  are lower than SwinIR~\cite{liang2021swinir} (See \tabref{tab:flops} for details).
It can be clearly seen that \namesrformer{}  achieves the best performance on almost all five benchmark datasets for all scale factors.
Since calculating self-attention within large windows can allow more information to be aggregated over a large area, our \namesrformer{}  performs much better on the high-resolution test set, such as Urban100 and Manga109.
Especially, for the $\times2$ SR training with DIV2K, our \namesrformer{}  achieves a 33.86dB PSNR score on the Urban100 dataset, which is 0.46dB higher than SwinIR but uses fewer parameters and computations.
The above strongly supports that our \namesrformer{}  is effective and efficient.
Through a series of enhancements to \namesrformer{} described in \secref{sec:srformerv2}, the upgraded \nameofmethod{} achieves better results, surpassing the previous state-of-the-art model HAT.

\newcommand{\addSubLightFigs}[2]{
	\renewcommand{\name}{figures/#1/}
	\begin{adjustbox}{valign=t}
		\begin{tabular}{c}
			\addImg{rectangle.jpg} \\
			#2
		\end{tabular}
	\end{adjustbox}
	\begin{adjustbox}{valign=t}
		\begin{tabular}{ccccc}
			\addImgS{HR}&
			\addImgS{Bicuble}&
			\addImgS{CARN}&
			\addImgS{IDN}&
			\addImgS{IMDN}
			\\
			HR &
			Bicubic &
			CARN~\cite{ahn2018fast} &
			IDN~\cite{hui2018fast} &
			IMDN~\cite{hui2019lightweight}
			\\
			\addImgS{EDSR-baseline}&
			\addImgS{LAPAR}&
			\addImgS{latticenet4x}&
			\addImgS{swinir-light} &		
			\addImgS{SRformer} 
			\\ 
			EDSR-baseline~\cite{lim2017enhanced}  &
			LAPAR-A~\cite{li2020lapar} &
			LatticeNet \cite{luo2020latticenet}&
			SwinIR-light~\cite{liang2021swinir}
			& \textbf{\namesrformer{}-light}
			\\
		\end{tabular}
	\end{adjustbox}
}

\begin{figure*}[thbp]
	\tablestyle{0.7pt}{1}
 	\scriptsize{
	\newcommand{\name}{figures/lightSR/Urban24/}
	\addSubLightFigs{lightSR/urban24}{Urban100 ($4\times$): img\_024} \\
	\addSubLightFigs{lightSR/urban67}{Urban100 ($4\times$): img\_067} \\
	\addSubLightFigs{lightSR/B100_92}{B100 ($4\times$): img\_78004} \\}
	\caption{Qualitative comparison of our \nameofmethod{-light} with recent 
	  state-of-the-art \textbf{lightweight image SR} methods 
		for the $\times 4$ SR task. 
		For each example, our \nameofmethod{-light} can restore the structures 
		and details better than other methods.
	}\label{fig:lightsr_visual}
\end{figure*}

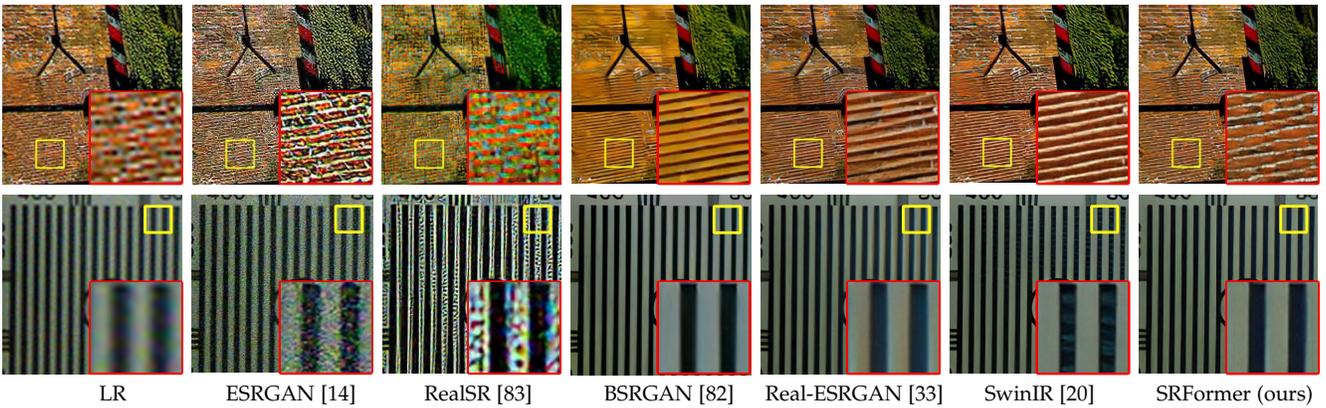
\begin{figure*}[!htbp]
    \centering
    \setlength{\abovecaptionskip}{0pt}
    \includegraphics[width=0.975\linewidth]{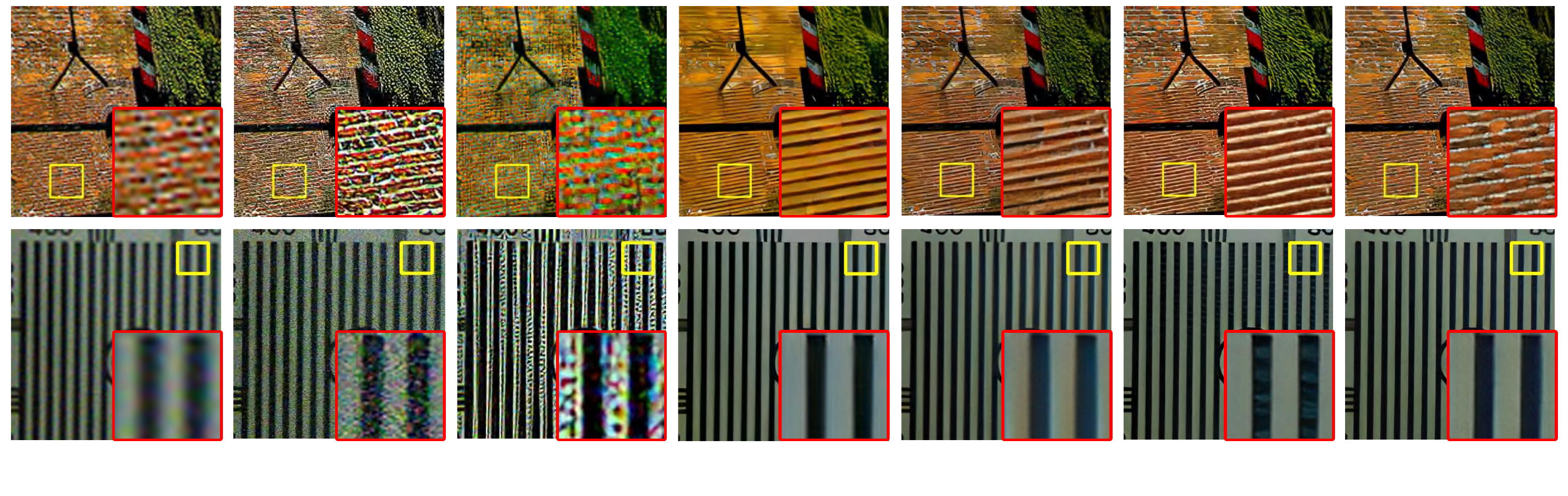}
    \footnotesize
    \put(-65., 8){\namesrformer{} (ours)}
    \put(-130., 8){SwinIR~\cite{liang2021swinir}}
    \put(-213., 8){Real-ESRGAN~\cite{wang2021real}}
    \put(-274., 8){BSRGAN~\cite{KaiZhang2021DesigningAP}}
    \put(-342., 8){RealSR~\cite{ji2020real}}
    \put(-417., 8){ESRGAN~\cite{wang2018esrgan}}
    \put(-465., 8){LR}
    \caption{Qualitative comparisons with recent state-of-the-art  methods on the $\times 4$ \textbf{real-world image SR} task. Our method can better restore the edges and textures compared to other methods.}
    \label{fig:real_sr}
\end{figure*}  

\myPara{Qualitative comparison.} We show qualitative comparisons of our \namesrformer{} and our advanced \nameofmethod{} with recent \sArt~methods in \figref{fig:sr_visual}. 
From each example of \figref{fig:sr_visual}, one can clearly observe that \namesrformer{}  can restore crisper and detailed textures and edges.
In contrast, the previous models restore blurred or low-quality textures. The qualitative comparison shows that our \namesrformer{} can restore better high-resolution images from the low-resolution ones.
Through a series of enhancements described in \secref{sec:srformerv2}, \nameofmethod{} further enhances image clarity and achieves much better results than all of the previous works.

\subsection{Lightweight Image Super-Resolution}
To demonstrate our model's scalability and further proof of\namesrformer{}'s efficiency and effectiveness, we train \namesrformer{-light} and compare it with a list of \sArt~lightweight SR methods: EDSR-baseline~\cite{lim2017enhanced}, CARN~\cite{ahn2018fast}, IMDN~\cite{hui2019lightweight}, LAPAR-A~\cite{li2020lapar}, LatticeNet~\cite{luo2020latticenet}, ESRT~\cite{lu2021efficient}, SwinIR-light~\cite{liang2021swinir}, and ELAN~\cite{zhang2022efficient}.

\myPara{Quantitative comparison.} The quantitative comparisons of lightweight image SR models are shown in \tabref{tab:light_weight_SR}. Following previous works~\cite{luo2020latticenet,ahn2018fast}, we report the MACs 
by upscaling a low-resolution image to  $1280 \times 720$ resolution on all scales.
We can see that our \namesrformer{}-light achieves the best performance on all five benchmark datasets for all scale factors. Our model outperforms SwinIR-light~\cite{liang2021swinir}  by up to $0.20$ dB PSNR scores on the Urban100 dataset and $0.25$ dB PSNR scores on the Manga109 dataset with even fewer parameters and MACs.
The results indicate that despite the simplicity, our permuted self-attention is a more effective way to encode spatial information.

\myPara{Qualitative comparison.} We compare our \namesrformer{} with \sArt~lightweight image SR models for qualitative comparisons  in \figref{fig:lightsr_visual}.
Notably, for all examples in \figref{fig:lightsr_visual}, \namesrformer{-light} is the only model that can restore the main structures with less blurring and artifacts.
This strongly demonstrates that the light version of \namesrformer{} also performs better for restoring edges and textures compared to other methods.

\subsection{Real-World Image Super-Resolution}

To test the performance of \namesrformer{} under a broader range of image degradation scenarios, we follow SwinIR~\cite{liang2021swinir} to retrain our \namesrformer{}  by using the same degradation model as BSRGAN~\cite{KaiZhang2021DesigningAP} and show results in \figref{fig:real_sr}.  \namesrformer{}  still produces more realistic and visually pleasing textures without artifacts when faced with real-world images, which demonstrates the robustness of our method. 

\subsection{Model Size Comparison}
While achieving the \sArt~performance, our \namesrformer{}  is also efficient. In \tabref{tab:flops}, we compare the parameters and MACs of our \namesrformer{}  with recent \sArt~models. Following previous works~\cite{luo2020latticenet,ahn2018fast}, we report the MACs 
by upscaling a low-resolution image to  $1280 \times 720$ resolution. One can clearly find that the performance of our \namesrformer{}  exceeds SwinIR~\cite{liang2021swinir} with fewer parameters and MACs. 

In our \nameofmethod{}, we explore the scaling capabilities of \namesrformer{}. Compared with HAT, our \nameofmethod{} has similar parameters and computational load but greatly exceeds HAT in performance.
We also train a smaller scaling model, named \nameofmethod{-S}, which follows the advanced operations in \figref{fig:roadmap} but has 180 channels. It requires much fewer computational resources than HAT but also performs better than it as shown in \tabref{tab:flops}. 
To further verify the effectiveness of PSA, we also replace the window attention in HAT with PSA using a $36 \times 36$ attention window size, which can also enhance the performance of HAT on both the Urban100 and Manga109 datasets.
\begin{table}[htp]
\setlength\tabcolsep{5.0pt}
\renewcommand{\arraystretch}{1}
\small
\centering
\setlength{\belowcaptionskip}{0cm}   
\caption{
Model size comparisons ($\times$2 SR). We report PSNR on Urban100 and Manga109. All models are trained on the DF2K (DIV2K~\cite{lim2017enhanced} + Flickr2K~\cite{timofte2017ntire}) training set.
}
\begin{tabular}{lcccc}
\toprule
Method &Params &MACs &   Urban100 &Manga109 \\ \midrule
SwinIR~\cite{liang2021swinir}  &11.80M &2.9T & 33.81 &39.92 \\ 
 SRFormer& 10.52M& 2.7T & 34.09 & 40.07 \\ \midrule
 HAT~\cite{chen2022activating}& 20.6M & 5.9T &  34.45 & 40.26 \\ 
 HAT + our PSA& 18.9M & 5.6T &  34.57 & 40.32 \\  \midrule
  \nameofmethod{-S}& 15.0M & 4.8T & 34.56 & 40.35\\
 \nameofmethod{}& 21.9M & 6.1T & \highlight{34.59} & \highlight{40.40}\\
				\bottomrule
\end{tabular}

\label{tab:flops}
\end{table}

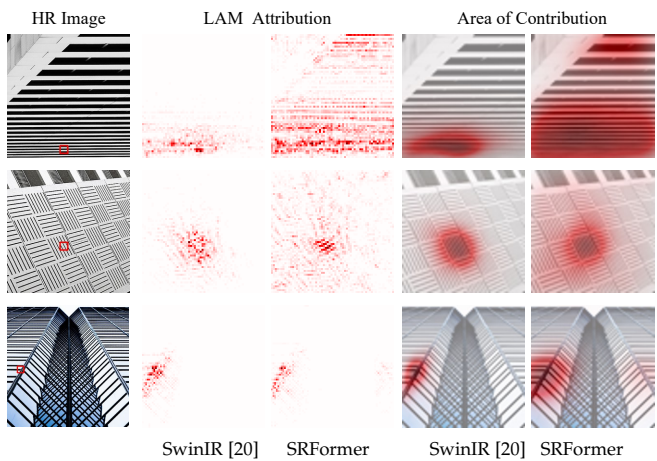
\begin{figure}[!htbp]
    \centering
    \scriptsize
    \includegraphics[width=\linewidth]{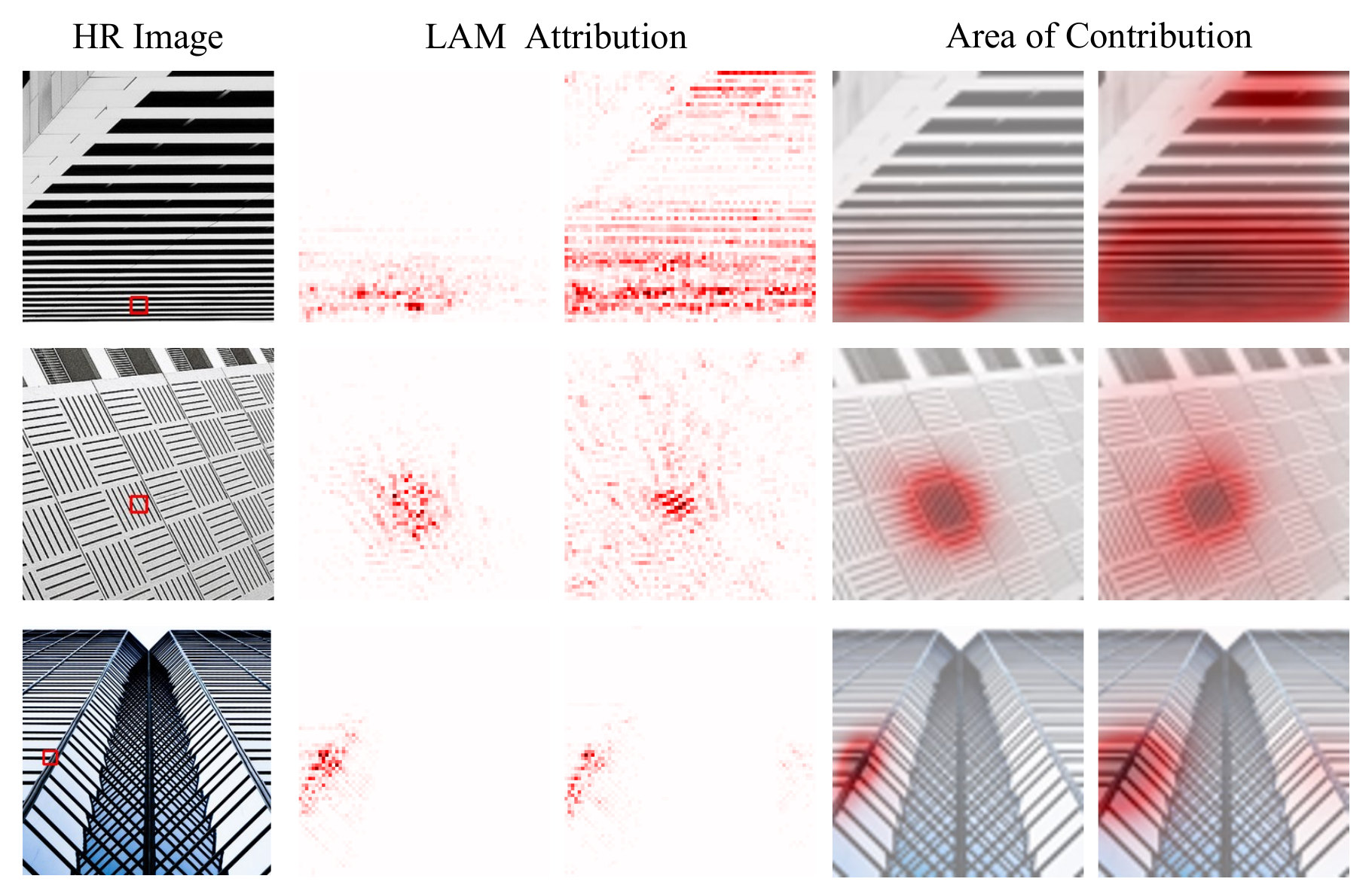}
    \put(-143, -8){\namesrformer{}}
    \put(-190, -8){SwinIR~\cite{liang2021swinir}}
    \put(-89, -8){SwinIR~\cite{liang2021swinir}}
    \put(-47, -8){\namesrformer{}}
    \caption{     LAM results of  SwinIR~\cite{liang2021swinir} and \namesrformer{} on  multiple challenging examples. We can see that \namesrformer{} can perform SR reconstruction based on a particularly wide range of pixels. }
    \label{fig:lam}
\end{figure}

\subsection{LAM Comparison}
To observe the range of utilized pixels for super-resolution (SR) reconstruction of our PSA, we compare \namesrformer{} with SwinIR using LAM~\cite{gu2021interpreting} on multiple challenging examples. 
Given a specified region in the SR image, LAM analyzes the contributions of each pixel in the input image to reconstruct this region. 
The results are shown in \figref{fig:lam}. From the first two examples, we can see that \namesrformer{}  performs reconstruction based on almost all the pixels of input because similar textures are repeated in the whole image. 
For the last example, a more amazing phenomenon can be observed that \namesrformer{} can span long distances and use the pixels on the right side for the reconstruction of the left region since the image is left-right symmetrical. 
This proves that the \namesrformer{} can extract the available features from almost the entire image for inference SR images in this case. 
Experimental results indicate that \namesrformer{} infers SR images with a significantly wider range of pixels than SwinIR~\cite{liang2021swinir}.

\section{Conclusion}

In this paper, we propose PSA, an efficient self-attention mechanism that can efficiently build pairwise correlations within large windows.
Based on our PSA, we design a simple yet effective Transformer-based model for single image super-resolution, called \namesrformer{}.
Due to the extremely large attention window and high-frequency information enhancement, \namesrformer{}  performs excellently on classical, lightweight, and real-world SR tasks.
In order to further investigate the potential of PSA, we further conduct scaling research and propose \nameofmethod{}, achieving new state-of-the-art performance.
We hope our permuted self-attention can be a paradigm of large window self-attention and serve as a useful tool for future research in super-resolution model design.

\ifCLASSOPTIONcaptionsoff
  \newpage
\fi

{\small
	\bibliographystyle{IEEEtran}
	\bibliography{egbib}
}

\newcommand{\addPhoto}[1]{\includegraphics[width=1in,height=1.15in,clip,keepaspectratio]{figures/bio/#1}}

\vspace{-35pt}
\begin{IEEEbiography}[\addPhoto{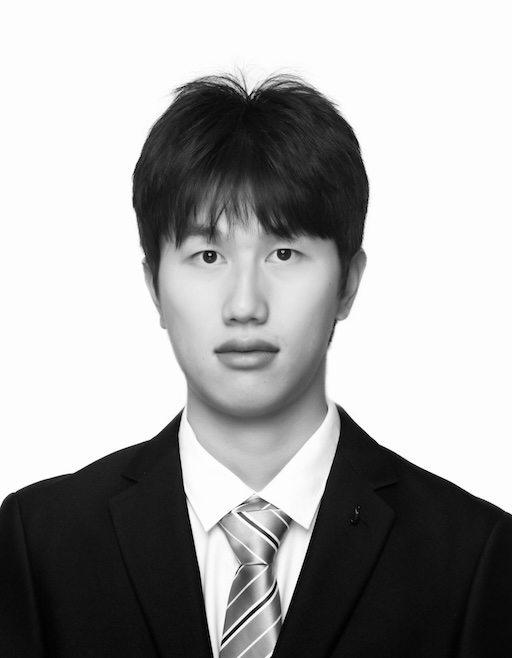}]{Yupeng Zhou} 
is currently working toward the PhD degree in the College of Computer Science, Nankai University, under the supervision of Prof. Qibin Hou. 
He received his bachelor's degree from Shandong University in 2022. 
His research interests include computer vision and deep learning, with a particular focus on image/video restoration and generation.
\end{IEEEbiography}

\vspace{-30pt}
\begin{IEEEbiography}[\addPhoto{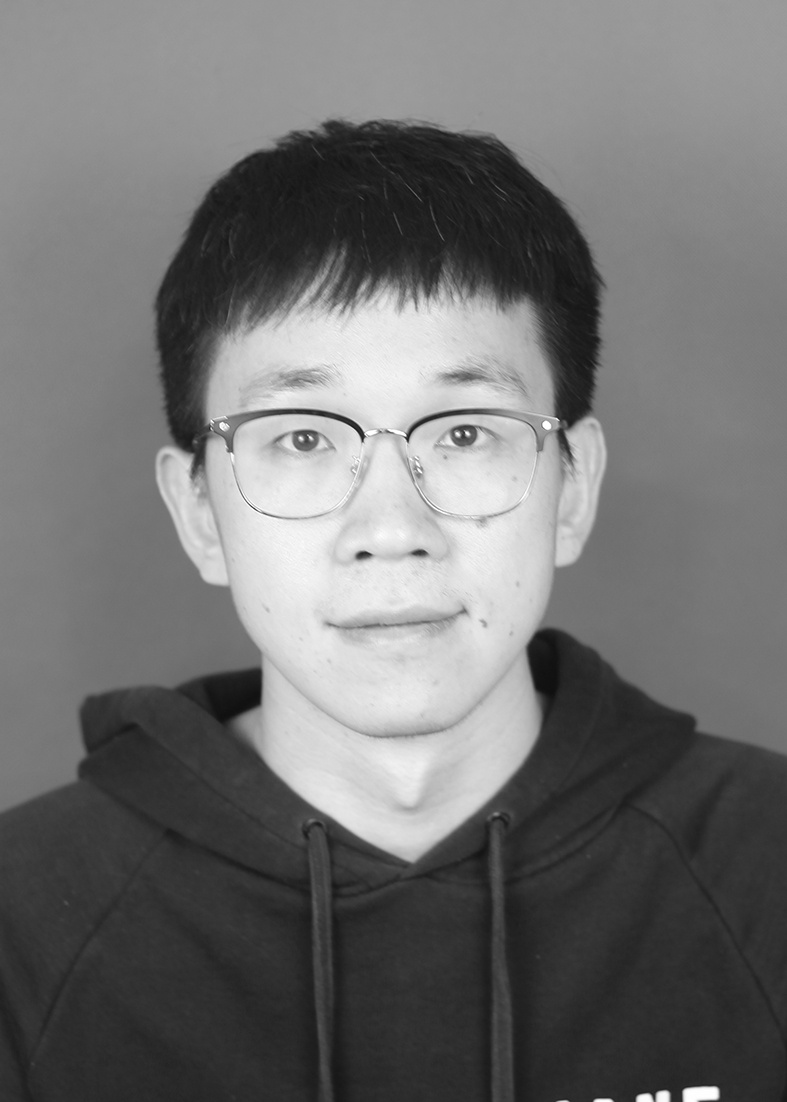}]{Zhen Li} 
is currently working toward the PhD degree in the College of Computer Science, Nankai University, under the co-supervision of Prof. Ming-Ming Cheng and Prof. Xiu-Li Shao. 
He received his MS degree from Sichuan University in 2019. 
His research interests include computer vision and deep learning, particularly focusing on image/video restoration and enhancement, generation and editing, etc.
\end{IEEEbiography}

\vspace{-30pt}
\begin{IEEEbiography}[\addPhoto{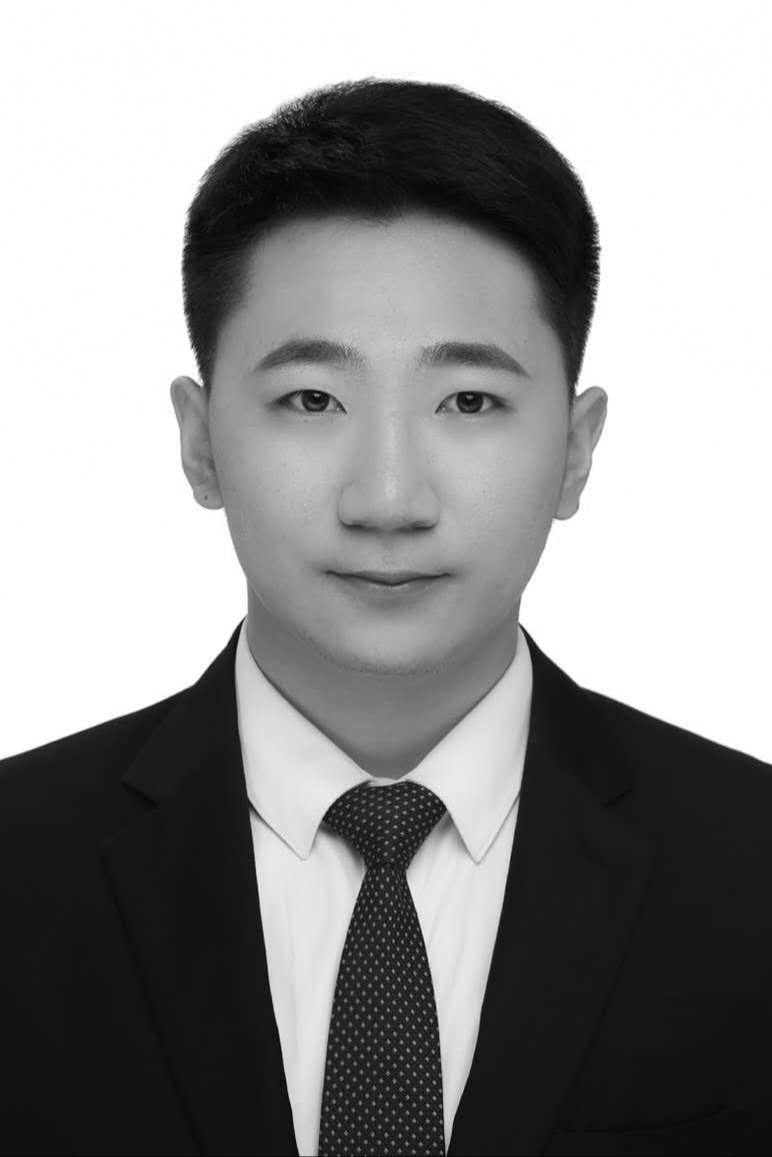}]{Chun-Le Guo}  received the Ph.D.degree from Tianjin University, China, under the supervision of Prof. ji-Chang Guo. He was a Visiting Ph.D. Student with the School of Electronic Engineering and Computer Science, Queen Mary University of London (QMUL), U.K. He was a Research Associate with the Department of Computer Science, City University of Hong Kong (CityU of HK). He was a Postdoctoral Researcher with Prof. Ming-Ming Cheng at Nankai University. He is currently an Associate Professor with Nankai University. His research interests include image processing, computer vision, and deep learning.
\end{IEEEbiography}

\vspace{-30pt}
\begin{IEEEbiography}[\addPhoto{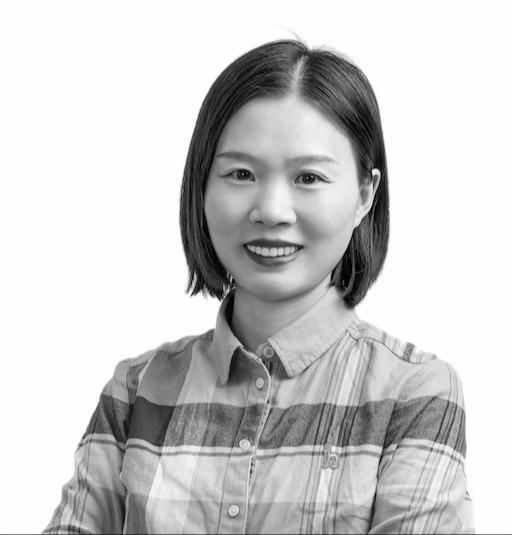}]{Li Liu} 
received her Ph.D. degree in information and communication engineering from the National University of Defence Technology, China, in 2012. She joined the faculty at the National University of Defense Technology in 2012. During her PhD study, she spent more than two years as a Visiting Student at the University of Waterloo, Canada, from 2008 to 2010. From 2015 to 2016, she spent ten months visiting the Multimedia Laboratory at the Chinese University of Hong Kong. From 2016 to 2018, she was a senior researcher of the CMVS at the University of Oulu, Finland. Dr. Liu was co-chair of nine International Workshops at several major venues, including CVPR, ICCV, and ECCV; she served as the leading guest editor of the special issues for IEEE TPAMI and IJCV. She also served as Area Chair for ICME 2020, 2021, 2022, and ACCV 2020, 2022. She currently serves as Associate Editor for IEEE Transactions on Circuits and Systems for Pattern Video Technology, IEEE Transactions on Geoscience and Remote Sensing, and Pattern Recognition. Her current research interests include computer vision, pattern recognition, and machine learning. Her papers currently have over 14000 citations in Google Scholar. She is a senior member of the IEEE.
\end{IEEEbiography}

\vspace{-30pt}
\begin{IEEEbiography}[\addPhoto{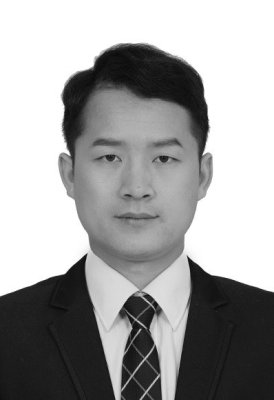}]{Ming-Ming Cheng} 
  received his PhD degree from Tsinghua University in 2012.
  Then he did 2 years research fellow, with Prof. Philip Torr in Oxford.
  He is now a professor at Nankai University, leading the Media Computing Lab.
  His research interests include computer graphics, computer vision, 
  and image processing. 
  He received research awards including 
  National Science Fund for Distinguished Young Scholars
  and ACM China Rising Star Award.
  He is on the editorial boards of IEEE TPAMI and IEEE TIP.
\end{IEEEbiography}

\vspace{-30pt}

\begin{IEEEbiography}[\addPhoto{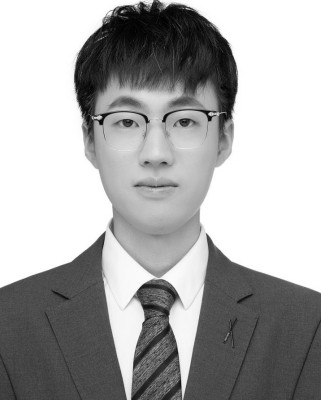}]{Qibin Hou} 
  received his Ph.D. degree from the School of Computer Science, 
  Nankai University. 
  Then, he worked at the National University of Singapore as a research fellow. 
  Now, he is an associate professor at School of Computer Science, 
  Nankai University. 
  He has published more than 30 papers on top conferences/journals, 
  including T-PAMI, CVPR, ICCV, NeurIPS, etc. 
  His research interests include deep learning and computer vision.
\end{IEEEbiography}
\vfill

\end{document}